\documentclass[journal]{IEEEtran}
\pdfoutput=1
\usepackage[nocompress]{cite}
\usepackage[pdftex]{graphicx}
\graphicspath{{./images/}{./biography/}{./figures/}}
\DeclareGraphicsExtensions{.pdf,.jpeg,.png}
\usepackage[cmex10]{amsmath}
\usepackage{algorithm}
\usepackage{algorithmic}
\usepackage[tight,normalsize,sf,SF]{subfigure}
\usepackage[cmex10]{amsmath}
\usepackage{amssymb}
\usepackage{stfloats}
\usepackage{amsthm}
\usepackage{bbm}
\usepackage{relsize}
\usepackage{booktabs}
\usepackage{multirow}
\usepackage{url}

\newcommand{\tensor}[1]{\boldsymbol{\mathcal{#1}}}

\newcommand{\normalpdf}[3]{\mathcal{N}\left(#1 \middle\vert#2 , #3\right)}

\newcommand{\hadamard}{\operatornamewithlimits{\mathlarger{\mathlarger{\mathlarger{\circledast}}}}}

\theoremstyle{plain}
\newtheorem{theorem}{Theorem}[section]
\newtheorem{lemma}{Lemma}[section]

\theoremstyle{definition}
\newtheorem{definition}[theorem]{Definition}
\theoremstyle{remark}

\begin{document}

\title{Bayesian Robust Tensor Factorization for \\Incomplete Multiway Data}

\author{Qibin Zhao,
        Guoxu Zhou,
        Liqing Zhang,
        Andrzej Cichocki,
        and Shun-ichi Amari
\thanks{Q. Zhao is with Laboratory for Advanced Brain Signal Processing, RIKEN Brain Science Institute, Japan and Department of Computer Science and Engineering, Shanghai Jiao Tong University, Shanghai, China.}
\thanks{ G. Zhou is with Laboratory for Advanced Brain Signal Processing, RIKEN Brain Science Institute, Japan.}
\thanks{ L. Zhang is with Key Laboratory of Shanghai Education Commission for Intelligent Interaction and Cognitive Engineering, Department of Computer Science and Engineering, Shanghai Jiao Tong University, Shanghai, China.}
\thanks{ A. Cichocki is with Laboratory for Advanced Brain Signal Processing,
RIKEN Brain Science Institute, Japan and Systems Research Institute in Polish Academy of Science,  Warsaw,  Poland.}
\thanks{ S. Amari is with Laboratory for Mathematical Neuroscience, RIKEN Brain Science Institute, Japan.}
}

\markboth{Journal of \LaTeX\ Class Files,~Vol.~13, No.~9, September~2014}%
{Shell \MakeLowercase{\textit{et al.}}: Bare Demo of IEEEtran.cls for Journals}

\maketitle

\begin{abstract}
We propose a generative model for robust tensor factorization in the presence of both missing data and outliers. The objective is to explicitly infer the underlying \emph{low-CP-rank} tensor capturing the global information and a sparse tensor capturing the local information (also considered as outliers), thus providing the robust predictive distribution over missing entries. The \emph{low-CP-rank} tensor is modeled by multilinear interactions between multiple latent factors on which the column sparsity is enforced by a hierarchical prior, while the sparse tensor is modeled by a hierarchical view of Student-$t$ distribution that associates an individual hyperparameter with each element independently. For model learning, we develop an efficient variational inference under a fully Bayesian treatment, which can effectively prevent the overfitting problem and scales linearly with data size. In contrast to existing related works, our method can perform model selection automatically and implicitly without need of tuning parameters. More specifically, it can discover the groundtruth of \emph{CP rank} and automatically adapt the sparsity inducing priors to various types of outliers.  In addition, the tradeoff between the low-rank approximation and the sparse representation can be optimized in the sense of maximum model evidence. The extensive experiments and comparisons with many state-of-the-art algorithms on both synthetic and real-world datasets demonstrate the superiorities of our method from several perspectives.
\end{abstract}

\begin{IEEEkeywords}
Tensor factorization, tensor completion, robust factorization, rank determination, variational Bayesian inference, video background modeling
\end{IEEEkeywords}

\IEEEpeerreviewmaketitle

\section{Introduction}
\IEEEPARstart{T}{ensors} (i.e., multiway arrays) can provide an efficient and faithful representation of structural properties for multidimensional data. For instance, a facial image ensemble affected by multiple conditions can be represented as a higher order tensor with dimensionality of $ pixel \times person \times pose \times illumination$. To model such data, tensor factorization has shown significant advantages in terms of capturing multiple interactions among a set of latent factors. Therefore its theory and algorithms have been an active area of study within the past decade, see e.g.~\cite{kolda2009tensor, Cichocki2009}, and have been successfully applied to various fields of applications such as face recognition, social network analysis, image and video completion, and brain signal processing~\cite{xiong2010temporal, gao2012probabilistic, zhao2013higher, chen2013simul}.  The two most popular tensor factorization frameworks are Tucker~\cite{de2000multilinear} and CANDECOMP/PARAFAC (CP)~\cite{kolda2009tensor}, also known as canonical polyadic decomposition (CPD)~\cite{sorensen2012canonical}, which naturally results in two different definitions of tensor rank, i.e., \emph{multilinear rank} and \emph{CP rank}.

When original data is only partially observed, tensor factorization can be applied for imputing the missing entries, known as tensor completion. CP factorization with missing values has been developed by employing CP weighted optimization (\mbox{CPWOPT})~\cite{acar2011scalable} and nonlinear least squares (CPNLS)~\cite{sorber2013optimization}.  To naturally deal with missing data, the probabilistic framework for tensor factorization was exploited~\cite{chu2009probabilistic, xiong2010temporal}, which has been extended to exponential family model~\cite{hayashi2010exponential} and nonparametric Bayesian model~\cite{qiinfinite}.  The main limitation of the existing tensor factorization scheme is that the tensor rank has to be specified manually, which tends to under- or over-fit the observations, resulting in severe deterioration of predictive performance. It is important to emphasize that our knowledge about the properties of tensor rank, especially \emph{CP rank}, is surprisingly limited~\cite{elizabeth2013tensor}. There is no straightforward algorithm to compute \emph{CP rank} of an explicitly given tensor, and the problem has been shown to be NP-hard~\cite{hillar2013most,de2008tensor}.  In fact, determining or even bounding the tensor rank is quite difficult in contrast to matrix rank~\cite{alexeev2011tensor, burgisser2011geometric}.
In~\cite{morup2009automatic}, ARD framework was applied to estimate the \emph{multilinear rank}. However, the solution is based on MAP point estimation and is not applicable to incomplete tensor data. Recently, Bayesian low-rank decomposition of incomplete tensors has been proposed in~\cite{rai2014scalable}, which can perform tensor completion while the CP rank can be also inferred by employing a multiplicative Gamma process prior. However, the missing data is handled by a heuristic way, i.e., estimating the missing data followed by the factorization on a whole tensor alternately. In addition, the inference is performed by Gibbs sampler which is generally shown slow convergence.

The convex optimization of nuclear norm has gained considerable attention in matrix completion, which essentially seeks the minimum rank under the condition of limited observations. Since \emph{multilinear rank} of a tensor is defined as the rank of its mode-$n$ matricizations, it can be optimized by simply applying nuclear norm based framework, yielding an extension to tensor completion~\cite{liu2013tensorcompletion}, which thus attracted many studies on low multilinear rank approximations~\cite{tan2014tensor,signoretto2013learning,huang2011composite}. In addition, the auxiliary information can be exploited to improve completion accuracy~\cite{narita2012tensor, chen2013simul}, which, however, is only suitable to some specific applications. It is worth noting that the convex optimization of nuclear norm requires several tuning parameters, which is prone to over- or under-estimate the tensor rank. In addition, since \emph{CP rank}, the standard definition of tensor rank, cannot be optimized by applying matrix techniques straightforwardly, its determination still remains challenging so far.

On the other hand, non-Gaussian noises or outliers may occur frequently in image and video data. To handle this problem, many robust techniques have been developed such as robust PCA~\cite{candes2011robust,xu2012robust,feng2013online,nie2011robust} and robust matrix factorization~\cite{xiong2011direct,zheng2012practical,nie2013joint,Eriksson2012Efficient,zhou2011godec}. For robust tensor factorization, 2DSVD using $R_1$-norm as the objective function was proposed by Huang et al.~\cite{huang2008robust}  and robust Tucker decompositions were  studied in~\cite{Zhang2013ICCV,li2011robust}. To handle both missing data and outliers, the nuclear norm regularization has been combined with $L_1$-norm loss function, which leads to a robust tensor completion~\cite{goldfarb2014robust}. However, the main limitation of above mentioned approaches is that the performance is quite sensitive to tuning parameters whose optimal selection is unrealistic or prohibitively expensive. For example, the parameter balancing model capacity between a low-rank term and a sparse term is generally tuned by performance evaluated on the groundtruth of missing data that is unknown in practice, implying that most existing approaches are impractical to obtain the optimal results. Therefore, an automatic model and parameter selection  based solely on observed data, which can achieve an optimal predictive performance, is appealing. Another limitation of existing robust tensor factorizations is that optimizations of latent factors are mainly based on point estimation, which is prone to overfitting especially when a large amount of missing data is present and not able to provide uncertainty information of predictions.

To address all these issues under a unified framework, we propose a probabilistic model with aim to recover the underlying low-rank tensor, modeled by multiplicative interactions among  multiple groups of latent factors, and an additive sparse tensor modeling  outliers, from partially observed data represented by a tensor of any order. More specifically, for the low-rank term, we specify a hierarchical sparsity-inducing  prior shared by multiple groups of latent factors, which gains a column sparsity along the latent components, resulting in an automatic determination of \emph{CP rank}. For the sparse term, a hierarchical view of Student-$t$ prior is placed independently on each element associated with an individual hyperparameter.  The top-level hyperparameters can be learned by maximizing  a lower-bound of the model evidence, resulting in that the sparsity constraint can be automatically adapted  to varying  fractions of outliers. To learn the model under a fully Bayesian framework, we derive a variational Bayesian algorithm for posterior inference, which is of high efficiency. Our method can be used for robust tensor factorization, robust tensor completion and anomaly detection with a significant advantage of automatic model and parameter selections without requiring any tuning parameters. Empirical results on both synthetic and real-world datasets demonstrate that the proposed method outperforms state-of-the-art methods in terms of predictive performance and robustness to outliers, even though the groundtruth is allowed to be used to tune the parameters in competing methods.

The rest of this paper is organized as follows. Section~\ref{sec:relatedwork} discusses the related work. Section~\ref{sec:preliminaries} introduces preliminary multilinear operations and notations.  Section~\ref{sec:BTF} presents  model specification and approximate Bayesian inference for robust tensor factorization, whose advantages are summarized in Section~\ref{sec:discussions}. Section~\ref{sec:results} shows extensive experimental results, followed by the conclusion in Section~\ref{sec:conclusion}.

\section{Related Work}
\label{sec:relatedwork}
Our work is somewhat related to probabilistic approaches for robust PCA~\cite{ding2011bayesian,luttinen2012bayesian} and for robust matrix factorization~\cite{wang2012probabilistic,wang2013bayesian,babacan2012sparse}. In \cite{ding2011bayesian}, Beta-Bernoulli distribution is exploited to model outliers and the low-rank matrix exclusively, which, however, results in high model complexity and slow inference. Missing values are considered in~\cite{luttinen2012bayesian}, where the number of latent components needs to be specified in advance. In~\cite{babacan2012sparse}, Jeffreys prior is adopted to model both noise and outliers. However, it cannot handle missing values. \mbox{PRMF}~\cite{wang2012probabilistic} uses Laplace distribution to model the residuals, while the rank of the underlying model should be given in advance. A fully Bayesian treatment of PRMF~\cite{wang2013bayesian} employs a hierarchical view of Laplace distribution as the noise model, and applies MCMC sampling for model inference. However, the rank also needs to be tuned manually and missing values are not considered. Finally, all these matrix based approaches cannot handle interactions of multiple factors, which is crucial for higher-order tensors.

Higher-order robust PCA (HORPCA), proposed very recently in~\cite{goldfarb2014robust}, is the only existing tensor method that can handle both missing data and outliers. It formulates the problem by a convex optimization framework in which nuclear norm and $L_1$-norm are exploited as regularization terms on the low-rank tensor and residual errors, respectively. However, it essentially optimizes the \emph{multilinear rank} and the predictive performance is sensitive to tuning parameters.  To our best knowledge, our paper is the first to present a fully Bayesian model for robust tensor factorization dealing with both missing data and outliers within one framework.

\section{Preliminaries and Notations}
\label{sec:preliminaries}
The order of a tensor is the number of dimensions, also known as ways or modes. Vectors are denoted by boldface lowercase letters, e.g., $\mathbf{a}$. Matrices are denoted by boldface capital letters, e.g., $\mathbf{A}$. Higher-order tensors (order $\geq 3$) are denoted by boldface calligraphic letters, e.g., $\tensor{A}$. Given an $N$th order tensor $\tensor{X}\in\mathbb{R}^{I_{1}\times I_{2}\times\cdots\times I_{N}}$, its $(i_1, i_2, \ldots, i_N)$th entry is denoted by $\mathcal{X}_{i_{1}i_{2}\ldots i_{N}}$ where the indices typically range from $1$ to their capital version, e.g., $i_{n}=1, 2,\ldots, I_{n}, n=1,\ldots, N$.

The inner product of two tensors is defined by $\langle \tensor{A}, \tensor{B}\rangle = \sum_{i_1i_2...i_N} \mathcal{A}_{i_1i_2...i_N}\mathcal{B}_{i_1i_2...i_N}$, and the squared Frobenius norm by $\|\tensor{A}\|_F^2 = \langle \tensor{A}, \tensor{A}\rangle$.
\begin{definition}
The \emph{generalized inner product} of $N\geq 3$  vectors, matrices, or tensors is defined as a sum of element-wise products. For example, \begin{equation}\label{eq:innerprodofmatrix}
\left\langle \mathbf{A}^{(1)}, \cdots, \mathbf{A}^{(N)} \right\rangle =\sum_{i,j} \prod_n A_{ij}^{(n)}.
\end{equation}
\end{definition}

The \emph{Hadamard product} is an entrywise product of two vectors, matrices or tensors of the same sizes. For instance,  $\mathbf{A}\in\mathbb{R}^{I\times J}$ and $\mathbf{B}\in\mathbb{R}^{I\times J}$, their Hadamard product, denoted by $\mathbf{A}\circledast\mathbf{B}$, is a matrix of size $I\times J$. Without loss of generality, the Hadamard product of a set of matrices $\{\mathbf{A}^{(n)}\}_{n=1}^{N}$ is simply denoted by
\begin{equation}
\hadamard_n \mathbf{A}^{(n)} = \mathbf{A}^{(1)}\circledast\mathbf{A}^{(2)}\circledast\cdots\circledast\mathbf{A}^{(N)}.
\end{equation}

The \emph{Kronecker product}\cite{kolda2009tensor} of matrices $\mathbf{A}\in\mathbb{R}^{I\times J}$ and $\mathbf{B}\in\mathbb{R}^{K\times L}$ is a matrix of size $IK \times JL$, denoted by $\mathbf{A}\otimes \mathbf{B}$. The \emph{Khatri-Rao product} of matrices $\mathbf{A}\in\mathbb{R}^{I\times K}$ and $\mathbf{B}\in\mathbb{R}^{J\times K}$ is a matrix of size $IJ\times K$ defined by a columnwise Kronecker product, and denoted by $\mathbf{A}\odot\mathbf{B}$. In particular, the Khatri-Rao product of a set of matrices in a reverse order is denoted by
\begin{equation}
\bigodot_{n}\mathbf{A}^{(n)} = \mathbf{A}^{(N)}\odot\mathbf{A}^{(N-1)}\odot\cdots\odot\mathbf{A}^{(1)},
\end{equation}
while the Khatri-Rao product of $\{\mathbf{A}^{(n)}\}_{n=1}^{N}$ except the $n$th matrix, denoted by $\mathbf{A}^{(\backslash n)}$,  is defined by
\begin{equation}
\small
\bigodot_{k\neq n}\mathbf{A}^{(k)} = \mathbf{A}^{(N)}\odot\cdots\odot\mathbf{A}^{(n+1)}\odot\mathbf{A}^{(n-1)}\odot\cdots\odot \mathbf{A}^{(1)}.
\end{equation}

\section{Bayesian Robust CP Factorization}
\label{sec:BTF}
\subsection{Model Specification}

\begin{figure}
\centering
  \includegraphics[width=0.45\textwidth]{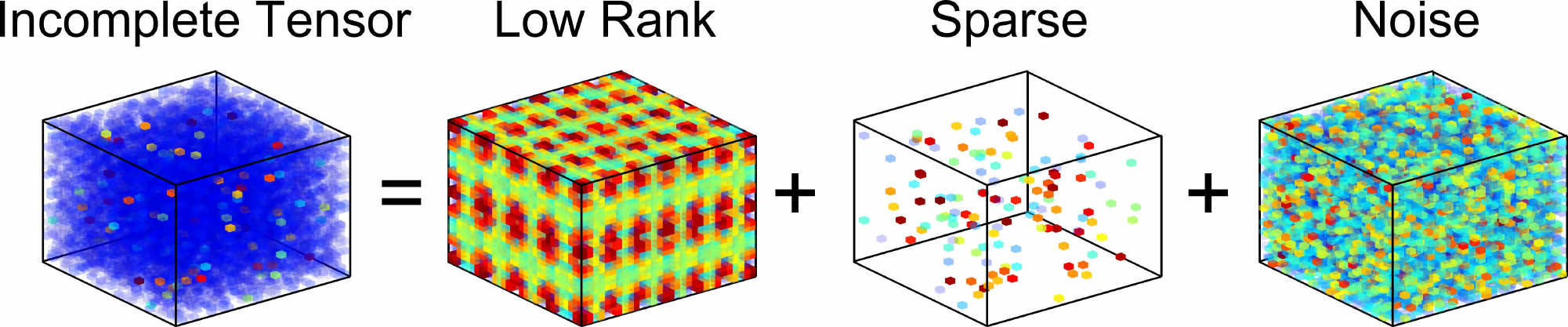}
  \caption{Bayesian robust tensor factorization. }
  \label{fig:concept}
\end{figure}

Let $\tensor{Y}$ be an incomplete $N$th-order tensor of size $I_1\times I_2\times \cdots \times I_N$ with missing entries. $\tensor{Y}_\Omega$ denotes the observed entries $\{\mathcal{Y}_{i_1 i_2 \ldots i_N}| ({i_1, i_2,\cdots, i_N}) \in \Omega\}$ where $\Omega$ denotes a set of indices.  We also define an indicator tensor $\tensor{O}$, whose entry $\mathcal{O}_{i_1 i_2\cdots i_N}$ is equal to 1 if $({i_1, i_2,\cdots, i_N}) \in \Omega$, otherwise is equal to $0$. We assume $\tensor{Y}$ is a noisy measurement of the true latent tensor $\tensor{X}$, and is corrupted by outliers $\tensor{S}$, i.e., $\tensor{Y} = \tensor{X} + \tensor{S} + \tensor{\varepsilon}$, where $\tensor{X}$ is generated by tensor factorization with a \emph{low-CP-rank}, representing the global information, $\tensor{S}$ is enforced to be sparse, representing the local information, and $\tensor{\varepsilon}$ is isotropic Gaussian noise (see Fig.~\ref{fig:concept}).

The standard CP factorization~\cite{kolda2009tensor} is expressed by
\begin{equation}\label{eq:CPmodel}
\tensor{X} = \sum_{r=1}^R \mathbf{a}^{(1)}_{\cdot r} \circ\cdots\circ \mathbf{a}^{(N)}_{\cdot r} = [\![ \mathbf{A}^{(1)},\ldots, \mathbf{A}^{(N)} ]\!],
\end{equation}
where $\circ$ denotes the outer product of vectors and $[\![\cdots]\!]$ is a shorthand notation of CP factorization. $\{\mathbf{A}^{(n)}|n=1,\ldots,N\}$ are latent factor matrices corresponding to each of $N$ modes, respectively. CP model can be interpreted as a sum of $R$ rank-one tensors, which is related to the definition of \emph{CP rank} that is the smallest integer $R$ for which the above representation holds. The \mbox{mode-$n$} factor matrix of size $I_n\times R$ can be denoted by row-wise or column-wise vectors, that is, $\mathbf{A}^{(n)} =\left[\mathbf{a}^{(n)}_{1},\ldots,\mathbf{a}^{(n)}_{I_n}\right]^T =\left[\mathbf{a}^{(n)}_{\cdot 1},\ldots,\mathbf{a}^{(n)}_{\cdot R} \right].$

To formulate robust CP factorization under the probabilistic framework, a generative model is introduced based on model assumptions. Specifically, the observation model is expressed by
\begin{multline}\label{eq:observationModel}
p\Big(\tensor{Y}_{\Omega} \Big\vert \{\mathbf{A}^{(n)}\}_{n=1}^N, \tensor{S}_\Omega, \tau\Big) =  \prod_{i_1=1}^{I_1}\cdots\prod_{i_N=1}^{I_N}\\
\mathcal{N}\left(\mathcal{Y}_{i_1  \ldots i_N} \middle\vert \left\langle\mathbf{a}^{(1)}_{i_1},\cdots,\mathbf{a}^{(N)}_{i_N}\right\rangle + \mathcal{S}_{i_1 \ldots i_N}, \tau^{-1}\right)^{\mathcal{O}_{i_1\cdots i_N}},
\end{multline}
where $\tau$ denotes the noise precision, $\mathbf{a}_{i_n}^{(n)}$ denotes the $i_n$th row vector of $\mathbf{A}^{(n)}$, and $\tensor{S}$ only has values corresponding to observed locations. The likelihood model in ($\ref{eq:observationModel}$) indicates that  $\mathcal{Y}_{i_1\cdots i_N}$ is generated by multiple $R$-dimensional latent vectors $\left\{\mathbf{a}^{(n)}_{i_n}\middle\vert n=1,\ldots,N\right\}$, whereas each  $\mathbf{a}_{i_n}^{(n)}$ affects a set of observations, i.e., a subtensor whose mode-$n$ index is $i_n$. The essential difference between matrix factorization and  tensor factorization is that the generalized inner product of $N(\geq3)$ latent vectors allows us to capture multilinear interactions reflecting the intrinsic structural property of data, which however leads to much more difficulties in model learning.

In practice, \emph{CP rank}, i.e., the dimensionality of latent space denoted by $R$, is unknown and considered as a tuning parameter whose optimal selection is quite challenging especially in the presence of missing data. Since $R$ controls the model complexity, we actually seek an automatic model selection strategy that can infer the true \emph{CP rank} from partially observed data.  To achieve this, in contrast to rank minimization on $\tensor{X}$, we attempt to minimize the dimensionality of latent space, which corresponds to column-wise sparsity of factor matrices.
Hence, we employ a sparsity inducing prior over factor matrices by associating an individual  hyperparameter to each latent dimension. More specifically, a hierarchical prior is equally specified over $N$ factor matrices, which is expressed by
\begin{equation}\label{eq:priorA}
\begin{split}
p\big(\mathbf{A}^{(n)}\big\vert \boldsymbol\lambda \big) &= \prod_{i_n=1}^{I_n} \mathcal{N}\big(\mathbf{a}_{i_n}^{(n)} \big\vert \mathbf{0}, \boldsymbol\Lambda^{-1} \big), \, \forall n\in [1,N] \\
p(\boldsymbol\lambda) &= \prod_{r=1}^{R}\text{Ga}(\lambda_r \vert c_0, d_0),
\end{split}
\end{equation}
where $\boldsymbol\Lambda=\text{diag}(\boldsymbol\lambda)$ denotes an inverse covariance matrix and is shared by latent factor matrices in all modes.
The hyperprior over $\boldsymbol\lambda$ is an i.i.d. Gamma distribution $ \text{Ga}(x\vert a,b) = \frac{b^a x^{a-1} e^{-bx}}{\Gamma(a)} $ where $\Gamma(a)$ is the Gamma function.

Due to the sparsity property, the initialization of R is usually set to its maximum possible value, while the effective dimensionality can be inferred automatically under Bayesian inference framework. For instance, if a particular $\lambda_r$ has a posterior distribution concentrated at large values, $\big\{\mathbf{a}^{(n)}_r| \forall n\in[1,N]\}$ will tend to be zero. Since the priors are shared by $N$ factor matrices, our framework can learn the same sparsity pattern for all factor matrices, yielding the minimum number of rank-one tensors.

The sparse term $\tensor{S}$ is modeled also by a hierarchical sparsity inducing prior. More specifically, Gaussian priors are placed on each data entry associated with an individual precision hyperparameter on which an i.i.d. Gamma hyperprior is placed, that is
\begin{equation}\label{eq:priorS}
\begin{split}
p(\tensor{S}_\Omega\vert\boldsymbol\gamma) &= \prod_{i_1,\ldots,i_N}\mathcal{N}(\mathcal{S}_{i_1 \ldots i_N}\vert 0, \gamma_{i_1 \ldots i_N}^{-1})^{\mathcal{O}_{i_1 \ldots i_N}},\\
p(\boldsymbol\gamma) &= \prod_{i_1,\ldots,i_N}\text{Ga}(\gamma_{i_1 \ldots i_N}\vert a_0^\gamma, b_0^\gamma).
\end{split}
\end{equation}
Note that when an individual parameter $\gamma_{i_1 \ldots i_N}$ goes to infinity, the corresponding element in $\tensor{S}$ is enforced to be exact zero.

The priors in (\ref{eq:priorA}) and (\ref{eq:priorS}) are related to the framework of sparse Bayesian learning (SBL)~\cite{tipping2001sparse} which is usually employed for variable selections. Since Laplacian and Student-$t$ distributions are commonly applied to enforcing sparsity, we may question why the choice of a Gaussian prior should express any preference for sparsity. In fact, (\ref{eq:priorS}) can be interpreted as  an infinite zero-mean Gaussian mixture with  mixture coefficients drawn from a Gamma distribution, which is thus a hierarchical view of Student-$t$ distribution. In other words, the marginal prior of $\tensor{S}_\Omega$ is an i.i.d. Student-$t$ distribution with the sparsity controlled by $(a_0^\gamma, b_0^\gamma)$ to some extent. For the case of noninformative hyperprior with $a_0^\gamma= b_0^\gamma=0$, we obtain the improper marginal prior $p(\mathcal{S}_{i_1 \ldots i_N})\propto 1/|\mathcal{S}_{i_1\ldots i_N}|$. Note that if $a_0^\gamma=1$, the hyperprior becomes an exponential distribution, such that the marginal prior over $\tensor{S}_\Omega$ is a Laplacian distribution. The elegance of this strategy therefore lies in the use of hierarchical modeling to obtain a prior which encourages sparsity while keeping fully conjugate exponential-family distributions throughout, which leads to the possibility of the fully Bayesian treatment.  Although our setting is related to SBL, the crucial difference lies in that our model specification can achieve column-wise sparsity, and the statistical property is shared by a set of factor matrices $\{\mathbf{A}^{(n)}\}_{n=1}^N$.

\begin{figure}
\centering
  \includegraphics[width=0.35\textwidth]{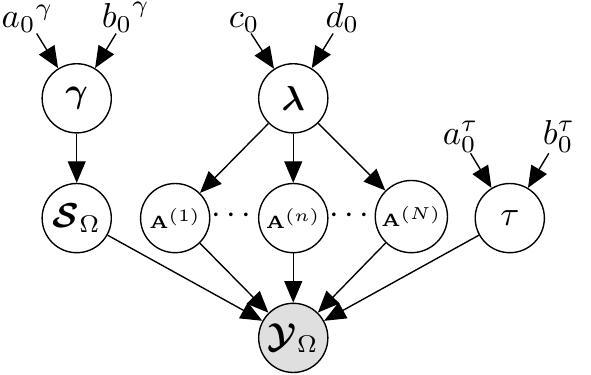}\\
  \caption{The probabilistic graphical model of Bayesian robust CP factorization of an incomplete tensor. }
  \label{fig:graphmodel}
\end{figure}

To complete the model, we also place a hyperprior over the noise precision $\tau$, that is
\begin{equation}\label{eq:hyperprior2}
p(\tau) = \text{Ga}(\tau\vert a_0^\tau, b_0^\tau).
\end{equation}
Finally, the probabilistic graphical model of robust tensor factorization is illustrated in Fig.~\ref{fig:graphmodel}. For simplicity of notations, all unknowns including both latent factor matrices and hyperparameters are collected and denoted together by $\Theta =\{\mathbf{A}^{(1)},\ldots,\mathbf{A}^{(N)}, \boldsymbol\lambda, \tensor{S}_\Omega, \boldsymbol\gamma, \tau \}$. Therefore, the joint distribution of the model, i.e., $p(\tensor{Y}_\Omega,\Theta)$, can be expressed by
\begin{equation}
\nonumber
\small
\begin{split}
\label{eq:jointdistribution}
p\left(\tensor{Y}_\Omega\middle\vert\{\mathbf{A}^{(n)}\}_{n=1}^N,\tensor{S}_\Omega,\tau\right)\prod_{n=1}^N p\big(\mathbf{A}^{(n)}\big\vert \boldsymbol\lambda \big) p(\tensor{S}_\Omega\vert\boldsymbol\gamma) p(\boldsymbol\lambda)p(\boldsymbol\gamma) p(\tau).
\end{split}
\end{equation}

In general, we can simply perform MAP estimation of $\Theta$ from the log-joint distribution (see Sec.~1 of Appendix) and most existing tensor factorization based on optimization approaches can be interpreted as point estimation by either maximum likelihood or MAP principles. However, in this study, we aim to provide a fully Bayesian treatment of the model by inferring the posterior distribution of $\Theta$, expressed by
$p(\Theta\vert\tensor{Y}_{\Omega})=  \frac{ p(\Theta,\tensor{Y}_\Omega)}{\int p(\Theta,\tensor{Y}_\Omega)\,d\Theta}.$
Thus the predictive distribution over missing entries $\tensor{Y}_{\backslash \Omega}$ can be also inferred by
$p(\tensor{Y}_{\backslash\Omega}\vert \tensor{Y}_{\Omega}) = \int  p(\tensor{Y}_{\backslash\Omega}\vert \Theta )p(\Theta\vert \tensor{Y}_\Omega)\, \text{d}\Theta.$

\subsection{Model Learning via Bayesian Inference}
Since exact Bayesian inference of our model is analytically intractable, we must resort to the approximate inference. Although variational Bayesian (VB) inference~\cite{winn2005variational} is difficult for derivations, especially when multiple interactions of latent factors are involved, it has advantages of closed-form posterior approximations and high efficiency as compared to sampling based inference methods. Hence, we employ VB inference to learn our model and present only the main results, while the detailed derivations and proofs are provided in Appendix\footnote{The Appendix is provided in supplementary materials.}.

We therefore seek a distribution $q(\Theta)$ to approximate the true posterior distribution $p(\Theta\vert \tensor{Y}_\Omega)$ in the sense of minimizing the KL divergence, that is
\begin{equation}\label{eq:VBKL}
\begin{split}
\small
\text{KL}\big(q(\Theta)\big\vert\big\vert p(\Theta\vert \tensor{Y}_\Omega)\big) &=\ln p(\tensor{Y}_\Omega) - \mathcal{L}(q), \\
 \mbox{where} \quad \mathcal{L}(q)  =&  \int q(\Theta)\ln\left\{\frac{p(\tensor{Y}_\Omega,\Theta)}{q(\Theta)}\right\}d\Theta.
\end{split}
\end{equation}
$\ln p(\tensor{Y}_\Omega)$ denotes the model evidence that is a constant, and its \emph{lower bound} is denoted by $\mathcal{L}(q)$. Thus, minimum of KL divergence implies the maximum of $\mathcal{L}(q)$. By applying mean field approximation, we assume that the posteriors can be factorized as
\begin{equation}
\label{eq:vbfactorization}
q\left(\Theta\right) = \prod_{n=1}^{N} q\left(\mathbf{A}^{(n)}\right) q(\tensor{S}_\Omega) q(\boldsymbol\lambda)q(\boldsymbol\gamma) q(\tau).
\end{equation}
Note that this is the only assumption about the distribution, while the particular functional forms of the individual factors can be explicitly derived in turn by virtue of conjugate exponential family in our hierarchical model.

\subsubsection{Posterior distribution of factor matrices}
From the graphical model shown in Fig.~\ref{fig:graphmodel}, the inference of mode-$n$ factor matrix $\mathbf{A}^{(n)}$ can be performed by receiving the messages from observed data, which are expressed by the likelihood term (\ref{eq:observationModel}), and incorporating the messages from their parents, which are expressed by the prior term (\ref{eq:priorA}).  The posteriors are shown to be factorized as independent distributions of their rows which are also Gaussian (see Sec. 2 of Appendix for details), i.e.,
\begin{equation}\label{eq:qDistributionA}
\begin{split}
q(\mathbf{A}^{(n)}) &= \prod_{i_n=1}^{I_n} \normalpdf{{\mathbf{a}}^{(n)}_{i_n} }{\tilde{\mathbf{a}}^{(n)}_{i_n} }{\mathbf{V}^{(n)}_{i_n} },\, \forall n\in [1,N],
\end{split}
\end{equation}
where the posterior parameters can be updated by
\begin{equation}\label{eq:updatefactor}
\small
\begin{split}
\tilde{\mathbf{a}}^{(n)}_{i_n} &= \mathbb{E}_q[\tau] \mathbf{V}^{(n)}_{i_n} \mathbb{E}_q\big[ \mathbf{A}_{i_n}^{(\backslash n)T}\big] \text{vec}\left(\tensor{Y}-\mathbb{E}_q[\tensor{S}]\right)_{\mathbb{I}(\mathcal{O}_{i_n}=1)}\\
\mathbf{V}^{(n)}_{i_n} &= \left(\mathbb{E}_q[\tau]  \mathbb{E}_q\big[\mathbf{A}_{i_n}^{(\backslash n)T}\mathbf{A}_{i_n}^{(\backslash n)}\big] + \mathbb{E}_q[\boldsymbol\Lambda] \right)^{-1}.
\end{split}
\end{equation}
$\mathbb{E}_q[\cdot]$ denotes the posterior expectation w.r.t. all variables involved. $\mathbb{I}(\mathcal{O}_{i_n}=1)$ is a sample function denoting a subset of the observed entries, whose mode-$n$ index is $i_n$. The most complex term in (\ref{eq:updatefactor}) is $\mathbf{A}_{i_n}^{(\backslash n)T} = \big(\bigodot_{k\neq n}\mathbf{A}^{(k)}\big)^T_{\mathbb{I}(\mathcal{O}_{i_n}=1)}$, where  $(\cdot)_{\mathbb{I}(\mathcal{O}_{i_n}=1)}$ denotes a subset of columns sampled according to the subtensor $\mathcal{O}_{i_n} =1$. Note that the update of $ \mathbf{V}^{(n)}_{i_n}$ involves expectation of the Khatri-Rao product, which can not be evaluated straightforwardly. Hence, we introduce the following results:

\begin{lemma}\label{lemma:1}
Given a set of independent random matrices $\{\mathbf{A}^{(n)}|n=1,\ldots,N\}$, we assume that $\forall n, \forall i_n$, the row vectors $\{\mathbf{a}^{(n)}_{i_n}\}$ are independent,  then
\begin{equation}\label{eq:expectationofcovariance}
\nonumber
\small
\mathbb{E}\left[ \Big(\bigodot_n \mathbf{A}^{(n)}\Big)^T \Big(\bigodot_n \mathbf{A}^{(n)}\Big)\right] =\sum_{i_1,\ldots,i_N} \hadamard_n \left(\mathbb{E}\left[ \mathbf{a}^{(n)}_{i_n}\mathbf{a}^{(n)T}_{i_n}\right] \right).
\end{equation}
\begin{proof}
See Sec.~3 of Appendix for details.
\end{proof}
\end{lemma}

According to Lemma \ref{lemma:1}, we can obtain that
\begin{equation}
\label{eq:ENATA}
\mathbb{E}_q\big[\mathbf{A}_{i_n}^{(\backslash n)T}\mathbf{A}_{i_n}^{(\backslash n)}\big] = \!\!\!\!\!\!\!\!\sum_{(i_1,\ldots,i_N)\in\Omega} \hadamard_{k\neq n} \left(\mathbb{E}\left[ \mathbf{a}^{(k)}_{i_k}\mathbf{a}^{(k)T}_{i_k}\right] \right).
\end{equation}
For simplicity,  let $\mathbf{B}^{(n)}$ of size $I_n\times R^2$ denote an expectation of a quadratic form related to $\mathbf{A}^{(n)}$ by defining $i_n$th-row vector $\mathbf{b}^{(n)}_{i_n} = \text{vec}\left(\mathbb{E}_q\left[ \mathbf{a}^{(n)}_{i_n}\mathbf{a}^{(n)T}_{i_n}\right]\right) = \text{vec}\left( \tilde{\mathbf{a}}^{(n)}_{i_n}\tilde{\mathbf{a}}^{(n)T}_{i_n} + \mathbf{V}^{(n)}_{i_n}\right)$,  then (\ref{eq:ENATA}) can be written as
\begin{equation}\label{eq:ATA}
\text{vec}\left(\mathbb{E}_q\big[\mathbf{A}_{i_n}^{(\backslash n)T}\mathbf{A}_{i_n}^{(\backslash n)}\big]\right)
= \Big(\bigodot_{k\neq n}\mathbf{B}^{(k)}\Big)^T \; \text{vec}(\tensor{O}_{\cdots i_n\cdots}).
\end{equation}
Note that the Khatri-Rao product in (\ref{eq:ATA}) is computed by all mode factors except $n$th mode, while the sum is performed according to the indices of observations, implying that only factors that interact with $\mathbf{a}^{(n)}_{i_n}$ are taken into account.

An intuitive interpretation of (\ref{eq:updatefactor}) is given as follows. $\mathbf{V}^{(n)}_{i_n}$ is updated by combining $\mathbb{E}_q[\boldsymbol\Lambda]$, denoting the factor prior, and covariance of other factor matrices computed by (\ref{eq:ATA}), while the tradeoff between these two terms is controlled by $\mathbb{E}_q[\tau]$ that is related to model fitness. In other words, the better fitness leads to more information from the current model than from the factor prior.
$\tilde{\mathbf{a}}^{(n)}_{i_n}$ is updated firstly by a linear combination of all other factors, while the combination coefficients are observed values, which implies that the larger observation leads to more similarity of its corresponding latent factors. Subsequently, $\tilde{\mathbf{a}}^{(n)}_{i_n}$ is rotated by $\mathbf{V}^{(n)}_{i_n}$ and is scaled according to the model fitness $\mathbb{E}_q[\tau]$.

\subsubsection{Posterior distribution of hyperparameters $\boldsymbol\lambda$}
From Fig.~\ref{fig:graphmodel}, the inference of $\boldsymbol\lambda$ can be performed by receiving messages from $N$ factor matrices and incorporating the messages from its hyperprior. We can show the posteriors of $\lambda_r,\forall r\in[1,R]$ are independent Gamma distribution, $q(\boldsymbol\lambda) = \prod_{r=1}^{R} \text{Ga}(\lambda_r \vert {c}_M^r, {d}_M^r )$,
where $c_M^r$, $d_M^r$ denote the posterior parameters learned from $M$ observations and can be updated by (see Sec.~4 of Appendix for details)
\begin{equation}\label{eq:lambda2}
c_M^r = c_0 + \frac{1}{2}\sum_{n=1}^{N} I_n, \quad
d_M^r = d_0 + \frac{1}{2}\sum_{n=1}^{N} \mathbb{E}_q\left[\mathbf{a}^{(n)T}_{\cdot r}\mathbf{a}^{(n)}_{\cdot r}\right].
\end{equation}
The posterior expectation term in (\ref{eq:lambda2}) can be evaluated using the posterior parameters in (\ref{eq:updatefactor}), thus we have $\mathbb{E}_q\left[\mathbf{a}^{(n)T}_{\cdot r}\mathbf{a}^{(n)}_{\cdot r}\right] = \tilde{\mathbf{a}}^{(n)T}_{\cdot r} \tilde{\mathbf{a}}^{(n)}_{\cdot r}  + \sum_{i_n} \left(\mathbf{V}_{i_n}^{(n)}\right)_{rr}$. Therefore, we can further simplify the computation of $\mathbf{d}_M=[d_M^1,\ldots d_M^R]^T$ by
\begin{equation}\label{eq:lambda3}
\mathbf{d}_M = \sum_{n=1}^{N} \left\{\text{diag}\left(\tilde{\mathbf{A}}^{(n)T}\tilde{\mathbf{A}}^{(n)} + \sum_{i_n} \mathbf{V}_{i_n}^{(n)} \right) \right\}.
\end{equation}
Based on the updated posterior of $\boldsymbol\lambda$, we can obtain $\mathbb{E}_q[\boldsymbol\Lambda] = \text{diag}([c_M^1/d_M^1,\ldots,c_M^R/d_M^R])$.

An intuitive interpretation is that the smaller $\sum_n \|\mathbf{a}^{(n)}_{\cdot r}\|_{2}^2$ leads to larger $\mathbb{E}_q[\lambda_r]$, which thus updates the prior over $\{\mathbf{a}^{(n)}_{\cdot r}\}_{n=1}^N$, resulting in that the $r$th component is enforced more strongly to be zero. Therefore, the smaller components can be diminished eventually to exact zero  and effectively pruned out after several iterations, while the larger components are enhanced to explain the data. This sparsity technique plays an key role to obtain the minimum number of components and automatic rank determination.

\subsubsection{Posterior distribution of sparse tensor $\tensor{S}$}
By combining the priors in (\ref{eq:priorS}) and the likelihood in (\ref{eq:observationModel}), we can derive the posterior approximation of $\tensor{S}$ as (see Sec.~5 of Appendix for details)
\begin{equation}
q(\tensor{S}) = \prod_{(i_1,\ldots,i_N)\in\Omega} \normalpdf{\mathcal{S}_{i_1 \ldots i_N} }{\tilde{\mathcal{S}}_{i_1 \ldots i_N}}{\sigma^2_{i_1 \ldots i_N}},
\end{equation}
where the posterior parameters can be updated by
\begin{equation}\label{eq:qDistS}
\begin{split}
\tilde{\mathcal{S}}_{i_1 \ldots i_N} &= \sigma^2_{i_1 \ldots i_N}\mathbb{E}_q[\tau]\left(\mathcal{Y}_{i_1\ldots i_N}-\mathbb{E}_q\left[\left\langle\mathbf{a}_{i_1}^{(1)},\ldots,\mathbf{a}_{i_N}^{(N)} \right\rangle \right] \right), \\
\sigma^2_{i_1 \ldots i_N} &= (\mathbb{E}_q[\gamma_{i_1\ldots i_N}] + \mathbb{E}_q[\tau])^{-1}.
\end{split}
\end{equation}
Observe that $\tensor{S}$ captures the information which is not explained by the low-rank CP approximation, while the magnitude is controlled by $\sigma_{i_1\ldots i_N}$ that is affected by the prior parameter $\mathbb{E}_q[\gamma_{i_1\ldots i_N}]$ and the precision of Gaussian noise $\mathbb{E}_q[\tau]$. The intuitive interpretation is that $\tensor{S}$ can model individual noises from total residuals, which are non-Gaussian. An alternative interpretation is that $ [\![ \mathbf{A}^{(1)},\ldots, \mathbf{A}^{(N)} ]\!]$ explains the global information by using minimum number of rank-one tensors, while $\tensor{S}$ explains the local information that is too expensive to be represented by increasing the model complexity.

\subsubsection{Posterior distribution of hyperparameters $\boldsymbol\gamma$}
By incorporating the prior and hyperprior of $\tensor{S}_\Omega$ in (\ref{eq:priorS}), we show that the posterior of $\boldsymbol\gamma$ is also factorized as independent distributions of each entries (see Sec.~6 of Appendix), given by
\begin{equation}
q(\boldsymbol\gamma) = \prod_{(i_1,\ldots,i_N)\in\Omega}\text{Ga}(\gamma_{i_1 \ldots i_N} \vert {a}_M^{\gamma_{i_1 \ldots i_N}}, {b}_M^{\gamma_{i_1 \ldots i_N}} ),
\end{equation}
whose posterior parameters can be updated by
\begin{equation}\label{eq:qDistGamma}
{a}_M^{\gamma_{i_1 \ldots i_N}} = a_0^\gamma + \frac{1}{2}, \,
{b}_M^{\gamma_{i_1 \ldots i_N}} = b_0^\gamma + \frac{1}{2}(\tilde{\mathcal{S}}^2_{i_1 \ldots i_N} + \sigma^2_{i_1 \ldots i_N}).
\end{equation}
This indicates that the smaller $\mathbb{E}_q[\mathcal{S}^2_{i_1 \ldots i_N}]$ leads to larger $\mathbb{E}_q[\gamma_{i_1 \ldots i_N}]$ which enforces $\tilde{\mathcal{S}}_{i_1 \ldots i_N}$ to be zero more strongly by (\ref{eq:qDistS}), and vice versa. In other words, the elements with small magnitude are forced to be zero, while the elements with large magnitude are further enhanced. It should be noted that the sparsity on $\tensor{S}$ is essentially important due to the fact that Gaussian with individual hyperparameters can easily capture the whole information of data.

\subsubsection{Posterior distribution of hyperparameter $\tau$}
The inference of noise precision $\tau$ can be performed by receiving the messages from observed data, and incorporating the messages from its hyperprior. We can show that the variational posterior is a Gamma distribution (see Sec.~7 of Appendix), i.e., $q(\tau) = \text{Ga}(\tau \vert {a}_M^\tau, b_M^\tau)$
where the posterior parameters can be updated by
\begin{equation}\label{eq:posteriorTau}
\small
\begin{split}
{a}_M^\tau &= a_0^\tau+\frac{1}{2}\sum_{i_1,\ldots,i_N}\mathcal{O}_{i_1 \ldots i_N},\\
{b}_M^\tau &= b_0^\tau+\frac{1}{2} \mathbb{E}_q\left[\left\|\tensor{O}\circledast\left(\tensor{Y}- [\![ \mathbf{A}^{(1)},\ldots, \mathbf{A}^{(N)} ]\!]-\tensor{S}\right)\right\|_F^2\right].
\end{split}
\end{equation}
However, the posterior expectation of model residuals in the above expression can not be computed straightforward, we need to introduce the following results.
\begin{lemma}\label{theorem:2}
Given a set of independent random matrices $\{\mathbf{A}^{(n)}|n=1,\ldots,N\}$, we assume that $\forall n, \forall i_n$, the row vectors $\{\mathbf{a}^{(n)}_{i_n}\}$ are independent,  then
\begin{multline}\label{eq:expX}
\mathbb{E}\left[ \left\| [\![ \mathbf{A}^{(1)},\ldots, \mathbf{A}^{(N)} ]\!]\right\|_F^2 \right]\\ = \sum_{i_1,\ldots,i_N} \left\langle \mathbb{E}\left[\mathbf{a}_{i_1}^{(1)}\mathbf{a}_{i_1}^{(1)T}\right],\ldots,\mathbb{E}\left[\mathbf{a}_{i_N}^{(N)}\mathbf{a}_{i_N}^{(N)T} \right]\right\rangle.
\end{multline}
\begin{proof}
See Sec.~8 of Appendix for details.
\end{proof}
\end{lemma}

By using Lemma \ref{theorem:2}, the posterior expectation in (\ref{eq:posteriorTau}) can be evaluated explicitly (see Sec.~9 of Appendix for details), that is
\begin{equation}\label{eq:modelerror}
\begin{split}
&\mathbb{E}_q\left[\left\|\tensor{O}\circledast\left(\tensor{Y}-[\![ \mathbf{A}^{(1)},\ldots, \mathbf{A}^{(N)}]\!]-\tensor{S}\right)\right\|_F^2\right]\\
=&\|\tensor{Y}_\Omega\|_F^2 - 2\text{vec}^T(\tensor{Y}_\Omega)\text{vec}\left([ \![ \tilde{\mathbf{A}}^{(1)},\ldots,\tilde{\mathbf{A}}^{(N)} ]\!]_\Omega \right) \\
&+ \text{vec}^T(\tensor{O}) \left(\bigodot_n \mathbf{B}^{(n)} \right)\mathbf{1}_{R^2} - 2\text{vec}^T(\tensor{Y}_\Omega)\text{vec}(\tilde{\tensor{S}}_\Omega) \\
& +2\text{vec}^T([\![\tilde{\mathbf{A}}^{(1)},\ldots,\tilde{\mathbf{A}}^{(N)} ]\!]_\Omega)\text{vec}(\tilde{\tensor{S}}_\Omega) + \mathbb{E}_q[\|\tensor{S}_\Omega\|_F^2].
\end{split}
\end{equation}
Hence, the posterior expectation of $\tau$ can be updated by $\mathbb{E}_q[\tau] = a_M^\tau/b_M^\tau$, where $a_M^\tau$ is related to the number of observations and $b_M^\tau$ is related to the posterior expectation of model residuals measured by squared Frobenius norm.

\subsubsection{Lower bound of model evidence}
We can also evaluate the variational lower bound in (\ref{eq:VBKL}) for our model. Since at each step of the iterative re-estimation procedure the value of this bound should not decrease, we can monitor the bound in order to test for convergence. The lower bound on the log marginal likelihood can be also written as
\begin{equation}\label{eq:lowerbound}
\mathcal{L}(q) = \mathbb{E}_{q(\Theta)}[\ln p(\tensor{Y}_\Omega,\Theta)] + H(q(\Theta)),
\end{equation}
where the first term denotes the posterior expectation of joint probability density, and the second term denotes the entropy of $q$ distribution. Taking the parametric form of $q$ distributions derived in the previous section, it can then be evaluated by an explicit form (See Sec.~10 of Appendix for details).

The top level hyperparameters $a_0^\gamma, b_0^\gamma, a_0^\tau, b_0^\tau,c_0, d_0$ are usually fixed to be very small values leading to a noninformative prior or set to zero leading to a Jeffrey's prior. Note that $a_0^\gamma, b_0^\gamma$ are related to sparsity degree, we seek a strategy to automatically adopt these hyperparameters to various types of outliers. This can be easily achieved by maximizing the lower bound w.r.t. $a_0^\gamma,b_0^\gamma$, expressed by
\begin{multline}\label{eq:updatetoplevel}
\mathcal{L}(a_0^\gamma,b_0^\gamma) = -M\ln\Gamma(a_0^\gamma) + Ma_0^\gamma \ln b_0^\gamma  \\ +(a_0^\gamma-1)\sum_{(i_1\ldots i_N)\in\Omega} \left\{(\psi(a_M^{\gamma_{i_1\ldots i_N}}) -\ln b_M^{\gamma_{i_1\ldots i_N}})\right\}\\
-b_0^\gamma  \sum_{(i_1\ldots i_N)\in\Omega}\
\frac{a_M^{\gamma_{i_1\ldots i_N}}}{b_M^{\gamma_{i_1\ldots i_N}}}.
\end{multline}

\subsubsection{Initialization of model inference}
The variational Bayesian inference is only guaranteed to converge to a local minimum. To alleviate getting stuck in poor local solutions, it is important to choose an initialization point. In our model, the top level hyperparameters including $c_0,d_0$, $a_0^\tau,b_0^\tau$, $a_0^\gamma, b_0^\gamma$ are set to $10^{-6}$, resulting in a noninformative prior. Thus the expectation of hyperparameters can be initialized by $\mathbb{E}[\boldsymbol\Lambda]=\mathbf{I}$, $\mathbb{E}[\tau]=1$ and $\forall n, \forall i_n,\mathbb{E}[\boldsymbol\gamma_{i_1\ldots i_N}] = 1$.  For the factor matrices, $\mathbb{E}[\mathbf{A}^{(n)}],\forall n\in[1,N]$ can be initialized by two different schemes. One is randomly drawn from $\mathcal{N}(\mathbf{0},\mathbf{I})$ for each row vector $\{\mathbf{a}^{(n)}_{i_n}\}$. The other is set to $\mathbf{A}^{(n)}= \mathbf{U}^{(n)}\boldsymbol\Sigma^{(n)^{\frac{1}{2}}}$, where $\mathbf{U}^{(n)}$ denotes the left singular vectors and $\boldsymbol\Sigma^{(n)}$ denotes the diagonal singular values matrix, obtained by SVD of \mbox{mode-$n$} matricization of $\tensor{Y}$. $\mathbf{V}^{(n)}$ is simply set to $\mathbb{E}[\boldsymbol\Lambda^{-1}]$. For sparse tensor $\tensor{S}$, $\mathbb{E}[\mathcal{S}_{i_1\ldots i_N}]$ is drawn from $\mathcal{N}(0,1)$, while $\sigma^2_{i_1\ldots i_N}$ is set to $\mathbb{E}[\boldsymbol\gamma^{-1}_{i_1\ldots i_N}]$.  The tensor rank $R$ is usually initialized by the maximum rank, i.e. $R\leq \min_n P_n$, where $P_n= \prod_{i\neq n} I_i$. For efficiency, we can also manually set the initialization value of $R$.

The whole procedure of model inference is summarized in Algorithm~\ref{alg:BRCPF}, where the posterior factors in (\ref{eq:vbfactorization}) are updated in an order that from bottom to top (see Fig.~\ref{fig:graphmodel}), which indicates that the message passing is started from observed data.

\begin{algorithm}[tb]
   \caption{Bayesian Robust Tensor Factorization }
   \label{alg:BRCPF}
\begin{algorithmic}
   \STATE {\bfseries Input:}  An $N$th-order incomplete tensor $\tensor{Y}$ and an indicator tensor $\tensor{O}$.
   \STATE {\bfseries Initialization:} $\tilde{\mathbf{A}}^{(n)},\mathbf{V}_{i_n}^{(n)}, \forall i_n\in[1,I_n],\forall n\in[1,N]$, $\tilde{\tensor{S}}, \sigma^2$,  hyperparameters $\boldsymbol\lambda, \boldsymbol\gamma, \tau$,  top level hyperparameters $c_0,d_0, a^\gamma_0,b^\gamma_0, a^\tau_0,b^\tau_0$.
   \REPEAT
   \FOR{$n=1$ {\bfseries to} $N$}
   \STATE Update the posterior $q(\mathbf{A}^{(n)})$ by (\ref{eq:updatefactor});
   \ENDFOR
   \STATE Update the posterior  $q(\boldsymbol\lambda)$ by (\ref{eq:lambda2});
   \STATE Update the posterior  $q(\tau)$ by (\ref{eq:posteriorTau});
   \STATE Update the posterior  $q(\tensor{S})$ by (\ref{eq:qDistS});
   \STATE Update the posterior  $q(\boldsymbol\gamma)$ by (\ref{eq:qDistGamma});
   \STATE Evaluate the lower bound by (\ref{eq:lowerbound});
   \STATE Update $a^\gamma_0,b^\gamma_0$ by maximizing (\ref{eq:updatetoplevel});
   \STATE Model reduction by eliminating zero-components in $\{\mathbf{A}^{(n)}\}$;
   \UNTIL{convergence.}
\end{algorithmic}
\end{algorithm}

\subsection{Predictive Distribution}
The predictive distribution over missing entries, given observed entries, is also analytically intractable. Hence, we can approximate it by using the variational posteriors of all parameters in $\Theta$, yielding a Student-$t$ distribution (see Sec.~11 of Appendix for details)
\begin{equation}\label{eq:predictivedistribution}
p(\tensor{Y} \vert \tensor{Y}_{\Omega}) = \prod\limits_{i_1,\ldots,i_N}\mathcal{T}(\mathcal{Y}_{i_1\ldots i_N} \vert \tilde{\mathcal{Y}}_{i_1 \ldots i_N}, \Psi_{i_1 \ldots i_N}, \nu_y)
\end{equation}
with its parameters given by
\begin{equation}
\nonumber
\small
\begin{split}
\tilde{\mathcal{Y}}_{i_1 \ldots i_N} &= \left\langle\tilde{\mathbf{a}}^{(1)}_{i_1},\cdots , \tilde{\mathbf{a}}^{(n)}_{i_N}\right\rangle,\\
 \Psi_{i_1\ldots i_N} &= \left\{\frac{b_M^\tau}{a_M^\tau} + \sum_n \left\{ \left(\hadamard_{k\neq n}\tilde{\mathbf{a}}_{i_k}^{(k)}\right)^T \mathbf{V}^{(n)}_{i_n} \left(\hadamard_{k\neq n}\tilde{\mathbf{a}}_{i_k}^{(k)}\right)\right\}\right\}^{-1},
\end{split}
\end{equation}
and $\nu_y = 2a_M^\tau$. Thus, the uncertainty of predictions can be obtained by $\text{var}(\mathcal{Y}_{i_1\ldots i_N}) = \frac{\nu_y}{\nu_y-2}\Psi_{i_1\ldots i_N}^{-1}$.

\subsection{Computational Complexity}
The computation cost of $N$ factor matrices in (\ref{eq:updatefactor}) is $O(R^2M \sum_n I_n + R^3 \sum_n I_n)$, where $M$ denotes the number of observations, $R$ denotes model complexity and is generally much smaller than the data size, i.e., $R\ll M$.   The computational costs are $O(R^2\sum_n I_n)$ for $\boldsymbol\lambda$, $O(R^2 M)$ for $\tau$, and $\mathcal{O}(MNR)$ for $\tensor{S}$.  Therefore, the overall complexity of our algorithm is $O( (R^2M + R^3) \sum_n I_n)$, which scales linearly with the data size but polynomially with the tensor rank. Note that due to the automatic model reduction, the excessive latent components are pruned out in the first few iterations such that $R$ reduces rapidly in practice. The solution presented in this paper mainly focuses on a general tensor factorization problem. However, when data is extremely sparse or large scale, an alternative strategy for approximate inference updated by each observed entry can be developed correspondingly.

\subsection{Case of Complete Tensor}
For fully observed tensor data, we can simply define $\tensor{O}$ with all elements being 1 and apply the model inference as described previously. However, several essentially different properties arise during inference, which leads to the possibility of more efficient computation for approximate posteriors. For the inference of factor matrices shown in (\ref{eq:updatefactor}), since $\mathbf{A}_{i_n}^{(\backslash n)}$ are same for any $i_n\in[1, I_n]$, such that $\{\mathbf{V}_{i_n}^{(n)}\}_{i_n=1}^{I_n}$ are all equivalent.  Hence, only one $\mathbf{V}^{(n)}$ needs to be computed for each mode-$n$, and $\{\tilde{\mathbf{a}}^{(n)}_{i_n}\}_{i_n=1}^{I_n}$ can be updated simultaneously by
\begin{equation}\label{eq:updatefactor2}
\begin{split}
\tilde{\mathbf{A}}^{(n)} &= \mathbb{E}_q[\tau]  \left(\mathbf{Y}_{(n)}-\mathbb{E}_q[\mathbf{S}_{(n)}]\right) \mathbb{E}_q\big[ \mathbf{A}^{(\backslash n)}\big]\mathbf{V}^{(n)},\\
\mathbf{V}^{(n)} &= \left(\mathbb{E}_q[\tau]  \mathbb{E}_q\big[\mathbf{A}^{(\backslash n)T}\mathbf{A}^{(\backslash n)}\big] + \mathbb{E}_q[\boldsymbol\Lambda] \right)^{-1},
\end{split}
\end{equation}
where $\mathbf{Y}_{(n)}$ denotes mode-$n$ matricization of tensor $\tensor{Y}$. For computational efficiency, we introduce another solution related to Lemma \ref{lemma:1}.

\begin{lemma}\label{lemma3}
Given a set of matrices $\{\mathbf{A}^{(n)}|n=1,\ldots,N\}$, if  the row vectors $\{\mathbf{a}^{(n)}_{i_n}\}_{i_n=1}^{I_n}$ are independent and $\text{cov}[\mathbf{a}^{(n)}_{i_n}] = \mathbf{V}^{(n)}, \forall i_n\in[1, I_n]$,  then
\begin{equation}
\nonumber
\mathbb{E}\left[ \Big(\bigodot_n \mathbf{A}^{(n)}\Big)^T \Big(\bigodot_n \mathbf{A}^{(n)}\Big)\right] =\hadamard_n \left(\mathbb{E}\left[ \mathbf{A}^{(n)T}\mathbf{A}^{(n)}\right] \right),
\end{equation}
where
$\mathbb{E}[\mathbf{A}^{(n)T}\mathbf{A}^{(n)}] = \left\{ \mathbb{E}\big[\mathbf{A}^{(n)T}\big]\mathbb{E}\big[\mathbf{A}^{(n)}\big] + I_n \mathbf{V}^{(n)}\right\}.$
\begin{proof}
See Sec.~12 of Appendix for details.
\end{proof}
\end{lemma}

According to Lemma \ref{lemma3}, the term in (\ref{eq:updatefactor2}) can be computed efficiently by
\begin{equation}
 \mathbb{E}_q\big[\mathbf{A}^{(\backslash n)T}\mathbf{A}^{(\backslash n)}\big] = \hadamard_{k\neq n}\left\{ \tilde{\mathbf{A}}^{(k)T}\tilde{\mathbf{A}}^{(k)} + I_k \mathbf{V}^{(k)}\right\}.
\end{equation}
Hence the computational cost for factor matrices are reduced to $\mathcal{O}(R^2\sum_n I_n + NR^3)$.  For hyperparameters $\boldsymbol\lambda$, the update rules in (\ref{eq:lambda3}) can be simplified by
\begin{equation}
\mathbf{d}_M = \sum_{n=1}^{N} \left\{\text{diag}\left(\tilde{\mathbf{A}}^{(n)T}\tilde{\mathbf{A}}^{(n)} + I_n \mathbf{V}^{(n)} \right) \right\}.
\end{equation}
The inference for $\tensor{S}$ and $\boldsymbol\gamma$ are similar to the case of incomplete data. For hyperparameter $\tau$, the posterior expectation of squared Frobenius norm of CP approximation can be computed more efficiently by introducing the following result.
\begin{lemma}\label{theorem:4}
Given a set of independent random matrices $\{\mathbf{A}^{(n)}|n=1,\ldots,N\}$, we assume that $\forall i_n\in[1, I_n]$, the row vectors $\{\mathbf{a}^{(n)}_{i_n}\}$ are independent,  then
\begin{multline}\nonumber
\mathbb{E} \left[ \left\| [\![ \mathbf{A}^{(1)},\ldots, \mathbf{A}^{(N)} ]\!]\right\|_F^2 \right] = \\
 \left\langle    \mathbb{E}\left[\mathbf{A}^{(1)T}\mathbf{A}^{(1)}\right],\ldots,\mathbb{E}\left[\mathbf{A}^{(N)T}\mathbf{A}^{(N)}\right]\right\rangle.
\end{multline}
\begin{proof}
See Sec.~13 of Appendix for details.
\end{proof}
\end{lemma}
Hence, the computational cost for $\tau$ is reduced to $\mathcal{O}(R^2\sum_n I_n)$. In addition, the computation of lower bound and predictive distributions can be also simplified easily, which would not been presented in details.

\section{Advantages}
\label{sec:discussions}
Since our model is based on a hierarchical probabilistic framework and fully Bayesian treatment, several advantages are gained and discussed as follows:
\begin{itemize}
  \item Our method is characterized as a \emph{tuning parameter free} approach and all model parameters can be learned automatically from observed data. By contrast, the existing tensor factorization methods require either predefined rank or penalty parameter and tensor completion methods using nuclear norm need to tune regularization parameters.
  \item The \emph{automatic rank determination} enables us to discover the ground-truth of CP rank, while the \emph{automatic sparsity model} can adapt the model to various types of outliers or non-Gaussian noises. Furthermore, the most elegant characteristic is that the tradeoff between the low-rank approximation and the sparse representation can be learned automatically in the sense of maximizing the model evidence.
  \item In contrast to point estimations by most existing tensor methods, the \emph{uncertainty information} over all model parameters are taken into account, which can effectively prevent overfitting problem. The full posteriors of factor matrices and predicted missing entries can provide confidence information regarding the solutions.
  \item An efficient and deterministic algorithm for Bayesian inference is developed, which empirically shows a \emph{fast convergence} and its computational complexity scales linearly with the data size.
\end{itemize}

\section{Experimental Results}
\label{sec:results}

\subsection{Validation on Synthetic Data}
We firstly assess the performance quantitatively on synthetic data. The true low-rank tensor $\tensor{X}$ of size $30\times 30\times 30$ was generated by rank-3 factor matrices, i.e., $\mathbf{A}^{(n)}\in\mathbb{R}^{30\times 3}, n=1,2,3$. Three components of the $n$th factor matrix are $[\sin(2\pi \frac{n}{30}i_n), \cos(2\pi \frac{n}{30}i_n), \mbox{sgn}(\sin(0.5  \pi i_n ))]$, indicating that the first two components possess different frequencies related to $n$, and the third components are common in all modes.  A random fraction of tensor entries were corrupted by outliers drawn from an uniform distribution $\mathcal{U}(-|H|,|H|)$. To mimic more realistic settings, a small noise drawn from $\mathcal{N}(0,0.01)$ was also considered. Subsequently, a fraction of entries were randomly selected to be observed tensor $\tensor{Y}_\Omega$, when missing data was considered. We utilized the root relative square error (RRSE), defined by $\frac{\|\hat{\tensor{X}}-\tensor{X}\|_2}{\|\tensor{X}\|_2}$, to evaluate the performance of tensor recovery. As for recovering the underlying factors, factor match error (FME)~\cite{sorber2013optimization} between the estimated factors and ground-truth was also evaluated. We compared our Bayesian robust tensor factorization (BRTF) with state-of-the-art methods including tensor factorization (CP-ALS~\cite{kolda2009tensor}, HOSVD~\cite{de2000multilinear}, CP-ARD~\cite{morup2009automatic}), tensor factorization with missing data (CPWOPT~\cite{acar2011scalable} and CPNLS~\cite{sorber2013optimization}), nonparametric Bayesian tensor factorization (MGPCP~\cite{rai2014scalable}) and robust tensor factorization with missing data (HORPCA~\cite{goldfarb2014robust}). It should be emphasized that the tuning parameters are necessary for most methods and have been carefully tuned based on ground-truth data, which is generally impractical for real applications. Specifically, CP-ALS, HOSVD, CPWOPT, CPNLS need to tune the parameter of \emph{tensor rank}, and HORPCA needs to tune the penalty parameter. By contrast, BRTF, MGPCP and CP-ARD can automatically estimate tensor rank and do not require any tuning parameters. Two scenarios were considered: 1) tensor recovery from fully observed data, and 2) tensor completion from partially observed data.

\begin{figure}[h]
\centering
\subfigure[\small $H = 10\cdot\text{std}(\text{vec}(\tensor{X}))$]{
   \includegraphics[width=0.95\columnwidth] {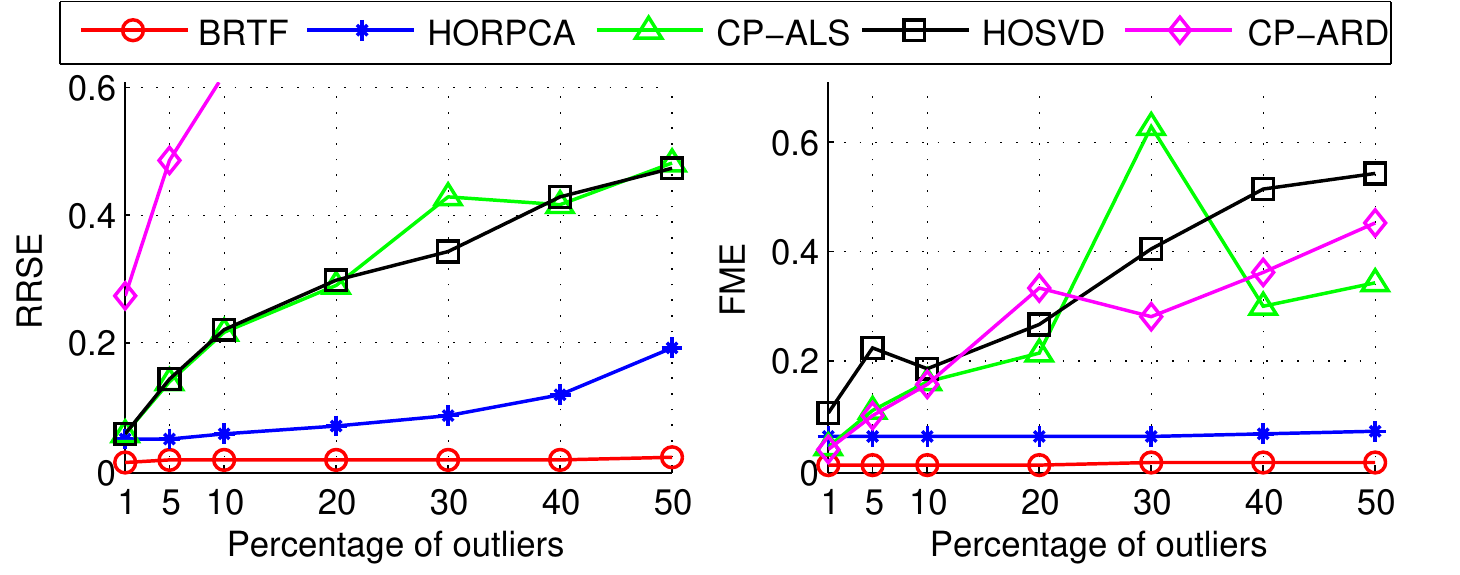}
   \label{fig:sim1}
 }
 \subfigure[\small $H= \max(\text{vec}(\tensor{X}))$]{
   \includegraphics[width=0.95\columnwidth] {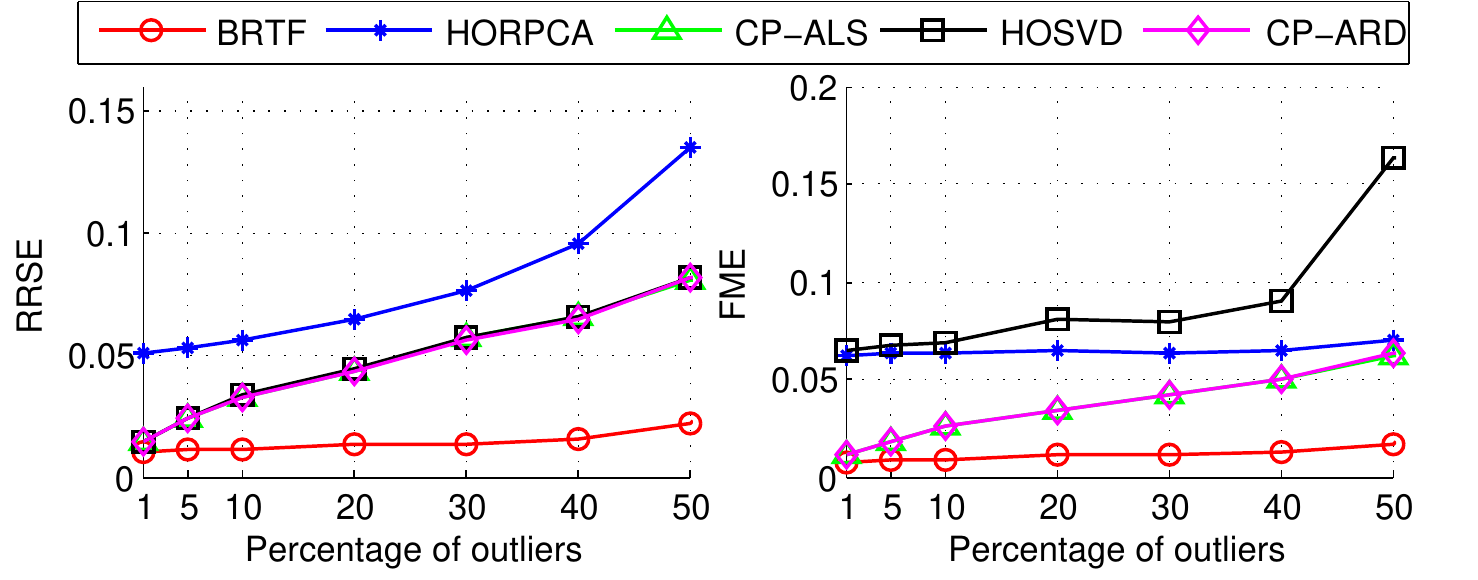}
   \label{fig:sim2}
   }
\caption{Performance of recovering true tensor $\tensor{X}$ and latent factors $\{\mathbf{A}^{(n)}\}_{n=1}^N$ under varying percentages of outliers. There are two types of outliers: \subref{fig:sim1} the magnitude is much larger than that of true signals, and  \subref{fig:sim2} is within the range of true signals.   }
\label{fig:simulation1}
\end{figure}

\begin{table}
\renewcommand{\arraystretch}{1.3}
\caption{\small Results on a complete tensor $\tensor{Y}$ of size $30\times 30\times 30$ with $R=50$ }
\label{tab:result1}
\centering
\resizebox{1\columnwidth}{!}
{
\begin{tabular}{ c | c | c | c | c | c}
 & BRTF & HORPCA & CP-ALS & HOSVD & CP-ARD \\
\hline
 Rank  &  50 (Auto) &  N/A  &  48 &   (23,23,23)  &  68 (Auto) \\
 RRSE &  0.04  & 0.20   &  0.31  &  0.56 &  0.35 \\
 FME &  0.02  & 0.96   &  0.29  &  0.96 &  0.20 \\
 Sensitivity &  N/A & 0.32  &  0.20 &   0.14  &  N/A \\
 Runtime &  3s & 10s  &  31s &   5s  &  4s \\
\end{tabular}
}
\vspace{-0.1in}
\end{table}

\begin{figure}[h]
\centering
\subfigure[\small $H = 10\cdot\text{std}(\text{vec}(\tensor{X}))$]{
   \includegraphics[width=0.95\columnwidth,height=1.3in] {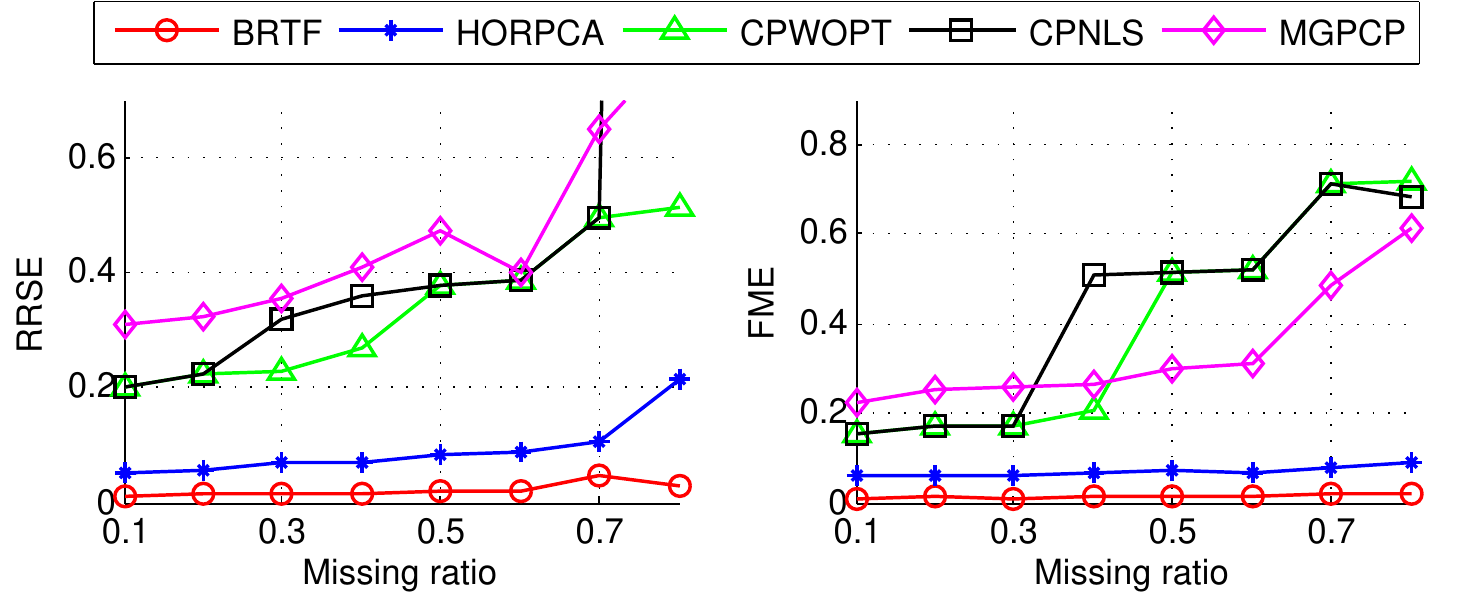}
   \label{fig:tc:sim1}
 }
 \subfigure[\small $H = \max(\text{vec}(\tensor{X}))$]{
   \includegraphics[width=0.95\columnwidth,height=1.3in] {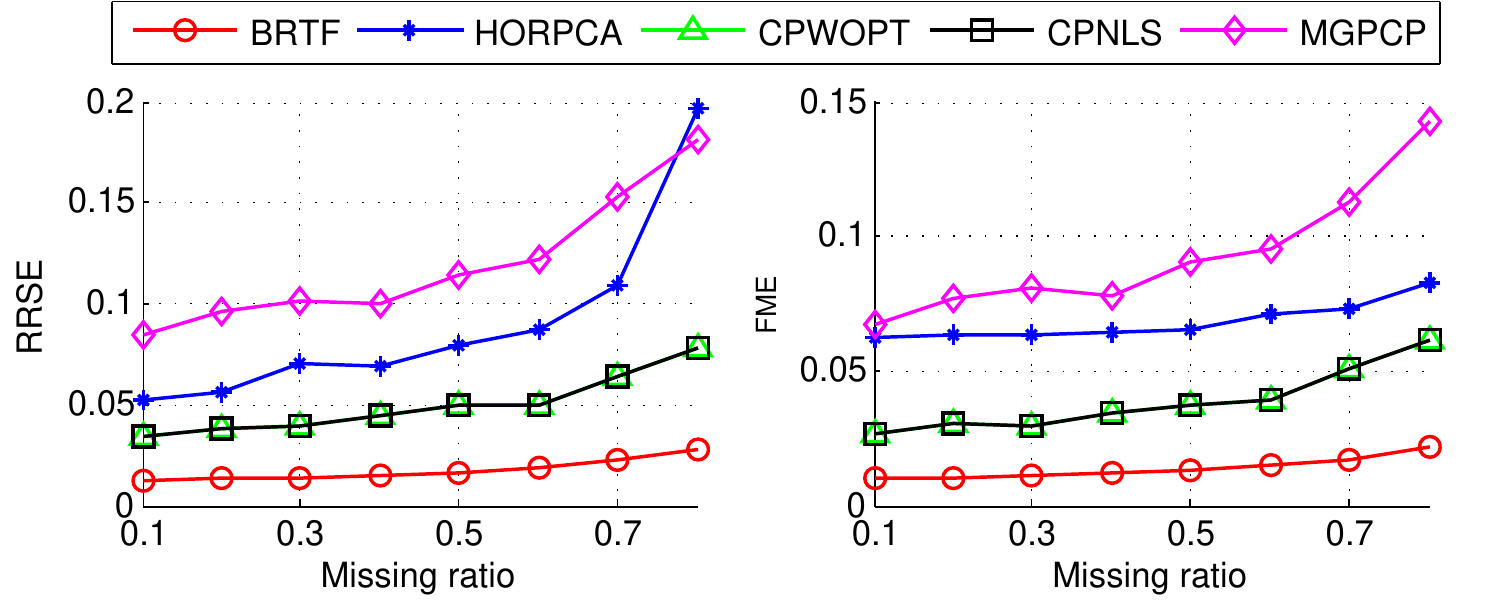}
   \label{fig:tc:sim2}
   }
\caption{Predictive performance on true tensor $\tensor{X}$ and latent factors $\{\mathbf{A}^{(n)}\}_{n=1}^N$ against fraction of missing entries. The percentage of outliers is fixed to 10\%, and two different magnitudes are considered in \subref{fig:tc:sim1}, \subref{fig:tc:sim2}. }
\label{fig:simulation2}
\end{figure}

\begin{table}[b]
\renewcommand{\arraystretch}{1.3}
\caption{\small The runtime (seconds) of simulations in Fig.~\ref{fig:tc:sim2} }
\label{tab:TimeCost}
\centering
%\resizebox{0.5\textwidth}{!}
{
\begin{tabular}{ c | c | c |c | c}
 BRTF & HORPCA & CPWOPT & CPNLS & MGPCP\\
\hline
2 $\pm$ 0.4 & 26 $\pm$ 13  &  26 $\pm$ 8 &  91 $\pm$ 10  & 68 $\pm$ 34 \\
\end{tabular}
}
\vspace{-0.1in}
\end{table}

\begin{figure*}
\centering
  \includegraphics[width=0.9\textwidth]{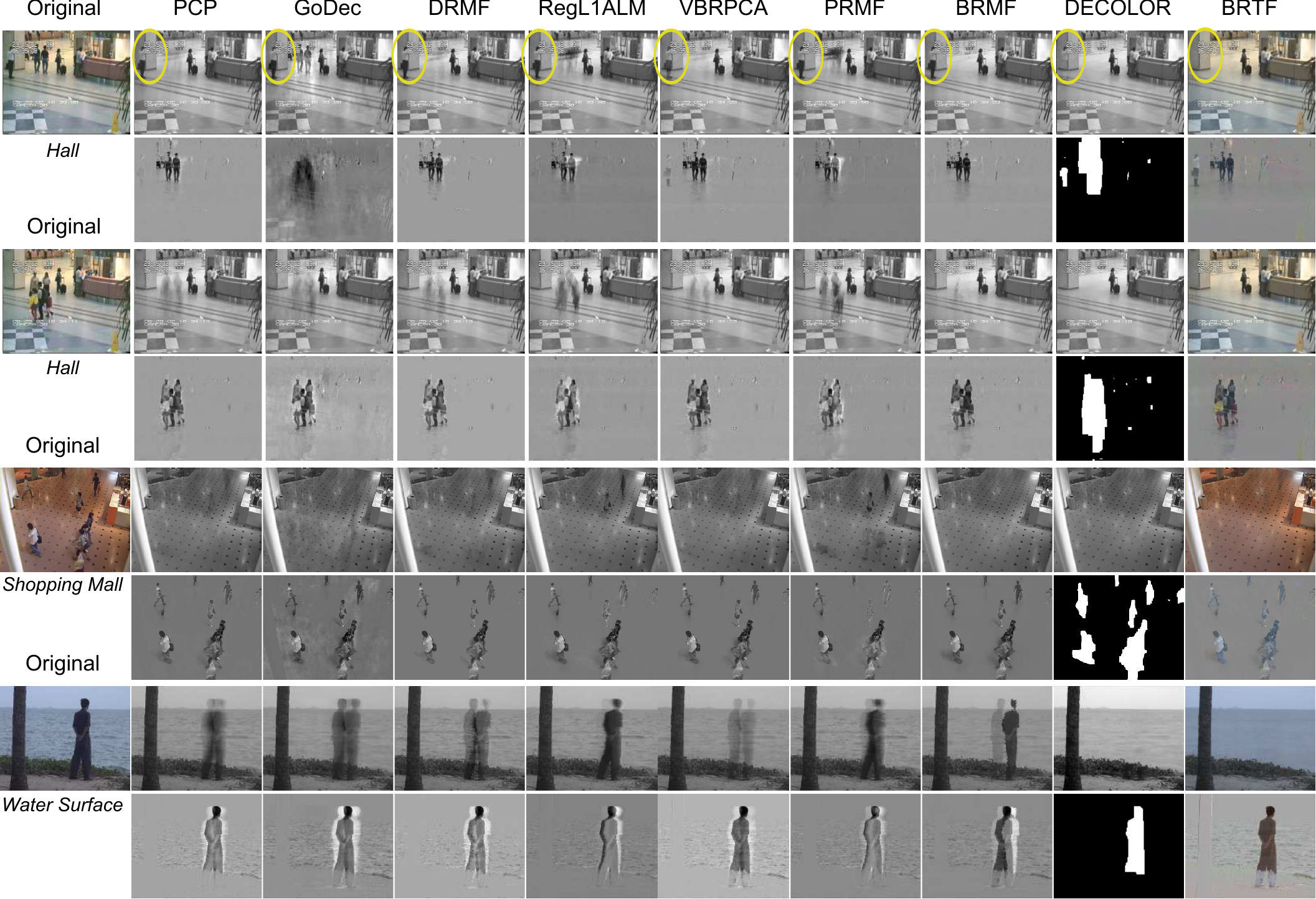}\\
  \caption{Results of background modeling. Four frames from three video sequences are shown from top to bottom. For each frame, there are two rows corresponding to background and foreground. From left to right, the results obtained by nine methods are shown in an order of the published years. The inferred CP ranks from three video sequences by BRTF are $2, 3, 3$ respectively. }
  \label{fig:vbm-complete}
\end{figure*}

The results for fully observed tensor are shown in Fig.~\ref{fig:simulation1}. Rank $R$ was initialized as $\max_n I_n$ for BRTF and CP-ARD. Observe that CPALS, CP-ARD and HOSVD are non-robust   due to the sensitivity of $L_2$-norm loss function to outliers. HORPCA is shown to be robust when outliers are much larger than true data, while it performs poorly when the magnitude of outliers is within the range of true data. It should be noted that BRTF significantly outperforms competing methods under all conditions in terms of recovering the low-rank tensor and latent factors, indicating that it is more robust to outliers and non-Gaussian noises. In addition, the performance of BRTF is unaffected by the percentage and magnitude of outliers, which confirms the capability of automatically adapting the model to various types of outliers.

A special case when the true \emph{CP Rank} is larger than data dimensions, i.e., $R>\max_n I_n$, was investigated under the condition of 1\% outliers and the detailed results are shown in Table~\ref{tab:result1}. The initial rank was set to 100 for both BRTF and CP-ARD. We observe that BRTF can correctly estimate rank $R$ while CP-ARD overestimates it due to the corruptions by outliers. For CP-ALS and HOSVD, the optimal rank was selected from all possible values within $[1,R]$ by multiple runs. The sensitivity of tuning parameters is also reported by standard deviation of RRSE under varying selections.  Although ground-truth data was used to tune parameters by other methods, BRTF still significantly outperforms all other methods in terms of RRSE and FME, while the runtime also shows its high efficiency. It should be noted that this experiment shows an essentially different property of tensor in contrast to matrix where $R\leq \min_n I_n$ is always satisfied. Therefore, the straightforward extension of many matrix based techniques is not applicable to this situation, which has been demonstrated by the low performance of HORPCA that employs the robust matrix technique to each mode-$n$ matricization of the tensor alternately.

The results for partially observed tensor are shown in Fig.~\ref{fig:simulation2}. Several tensor factorization based completion methods are compared under varying missing ratios. For computation efficiency, the rank R was initialized as 10 in BRTF and MGPCP.  Observe that HORPCA is more robust than MGPCP, CWOPT and CPNLS, which cannot handle outliers explicitly, only when outliers are much larger than true data. By contrast, BRTF achieves the best performance among competing methods and its performances are quite stable under varying missing ratios, which demonstrates its robustness to both outliers and missing data. In addition, the results confirm that BRTF can accurately estimate the ground-truth of tensor rank in all cases. It should be emphasized that all competing methods had multiple runs to tune the parameters and the best possible predictive performance on missing data was reported, while BRTF and MGPCP only needs to run once. These results demonstrate that the superiorities of BRTF is not only in automatic model selection, but also in the accuracy of tensor factorization and completion. In addition, Table~\ref{tab:TimeCost} shows the mean and standard deviation of runtime under different missing ratios, indicating the high efficiency of BRTF.

\subsection{Video Background Modeling}
Anomaly detection is another important ability of BRTF which can model the local information explicitly. Hence, we now consider a real-world application in surveillance video sequences with aim to separate the foreground objects from the background. Since the background is highly correlated along the frames, thus can be modeled by a low-rank tensor, while the foreground objects are moving along frames, thus can be modeled by a sparse tensor. We conducted experiments on the popular video sequences\footnote{\url{http://perception.i2r.a-star.edu.sg/bk_model/bk_index.html}} by extracting 100 frames from \emph{Shopping mall} sequence with frame size of $256\times 320$, 300 frames from \emph{Hall} sequence with frame size of $144\times 176$ and 300 frames from \emph{WaterSurface} sequence with frame size of $128\times 160$. The state-of-the-art methods for robust matrix/tensor factorization and background modeling were employed for comparisons. The matrix-based methods, including PCP~\cite{candes2011robust}, GoDec~\cite{zhou2011godec}, DRMF~\cite{xiong2011direct}, RegL1ALM~\cite{zheng2012practical}, VBRPCA~\cite{babacan2012sparse}, PRMF~\cite{wang2012probabilistic}, BRMF~\cite{wang2013bayesian}, and DECOLOR~\cite{zhou2013moving}, were performed on grayscale videos represented by matrices (e.g., $81920\times 100$ for \emph{Shopping mall}) while HORPCA~\cite{goldfarb2014robust} and BRTF were performed on original color videos represented by tensors with an additional RGB mode (e.g., $81920\times 3 \times 100 $ for \emph{Shopping mall}). For competing methods, if necessary, the tuning parameters were selected close to optimal by visual quality due to the lack of ground-truth. The initialized rank in BRTF is set to 10.

As shown in Fig.~\ref{fig:vbm-complete}, BRTF successfully separates the background and foreground on all the sequences. From one frame of \emph{Hall}, we observe that BRTF can completely separate the person, who stands for a while and then moves away, from the background, while all other methods capture this person by both background and foreground. In another frame of \emph{Hall}, the ghosting effects are observed by all methods except DECOLOR and BRTF. For the sequence of \emph{Shopping Mall}, VBRPCA, BRMF, DECOLOR and BRTF obtain clearer background than those by other methods. For the sequence of \emph{Water Surface}, DECOLOR and BRTF are clearly superior to other methods. Note that DECOLOR obtains comparable results with BRTF, since it incorporates the auxiliary information that is specially designed for this application. By contrast, BRTF, as a general tool for robust factorization, has shown the superiorities not only in handling tensor data, but also in more robustness than existing robust matrix factorizations. An intuitive interpretation is that BRTF can be considered as robust matrix factorizations on R, G, B matrices simultaneously with a constraint of common factors, which is more effective to capture the low-rank structure than applying matrix factorization independently.

\begin{figure}[tb]
\centering
  \includegraphics[width=0.45\textwidth]{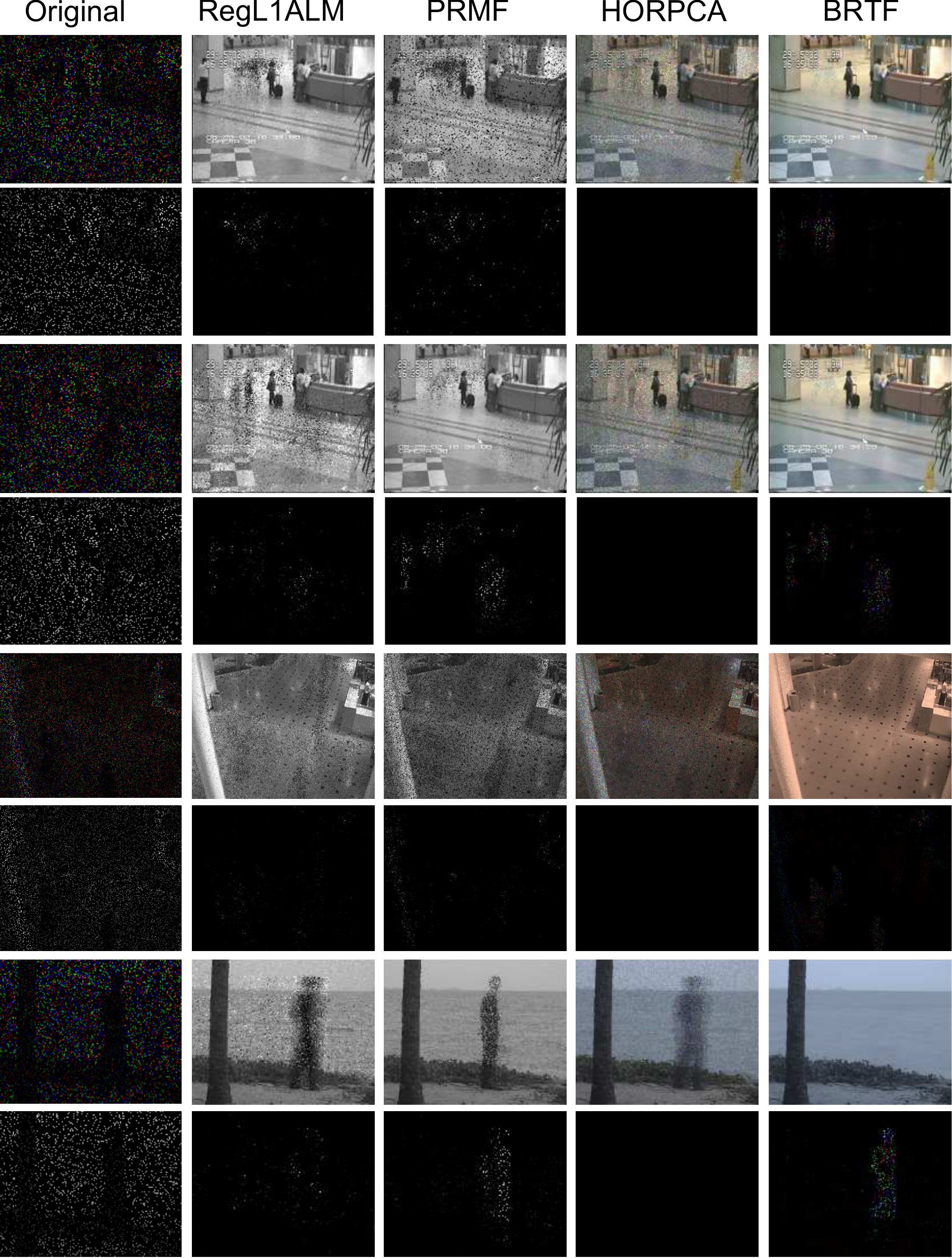}\\
  \caption{Background modeling when 90\% pixels are missing. Four frames from several sequences are shown and each frame has two rows corresponding to background and foreground. The original frame in color and grayscale forms are shown in the first column, followed by results obtained by different methods.  The inferred CP ranks from three video sequences by BRTF are $2, 1, 1$ respectively. }
  \label{fig:vbm-incomplete}
\end{figure}

To illustrate the property of simultaneous robust completion and anomaly detection, we conducted additional experiments on the same sequences by randomly dropping 90\% pixels and compared BRTF with RegL1ALM, PRMF and HORPCA, which can handle both missing data and outliers. As illustrated in Fig.~\ref{fig:vbm-incomplete}, BRTF is significantly superior to matrix based RegL1ALM and PRMF and tensor based HORPCA in terms of recovering the background. We observe that in the presence of missing pixels, the ghost effects are more severe by other methods, while BRTF is unaffected, indicating its better robustness to missing data and outliers. Although HORPCA is also a robust tensor factorization method, it shows comparable results with matrix based methods and cannot recover well the color of background. It should be noted that all competing methods require tuning parameters whose selection is quite time consuming, while BRTF works in a fully automatic fashion.

\subsection{Facial Image Denoising}
\begin{figure*}
\centering
\subfigure[\small Corrupted images]{
   \includegraphics[width=0.3\textwidth] {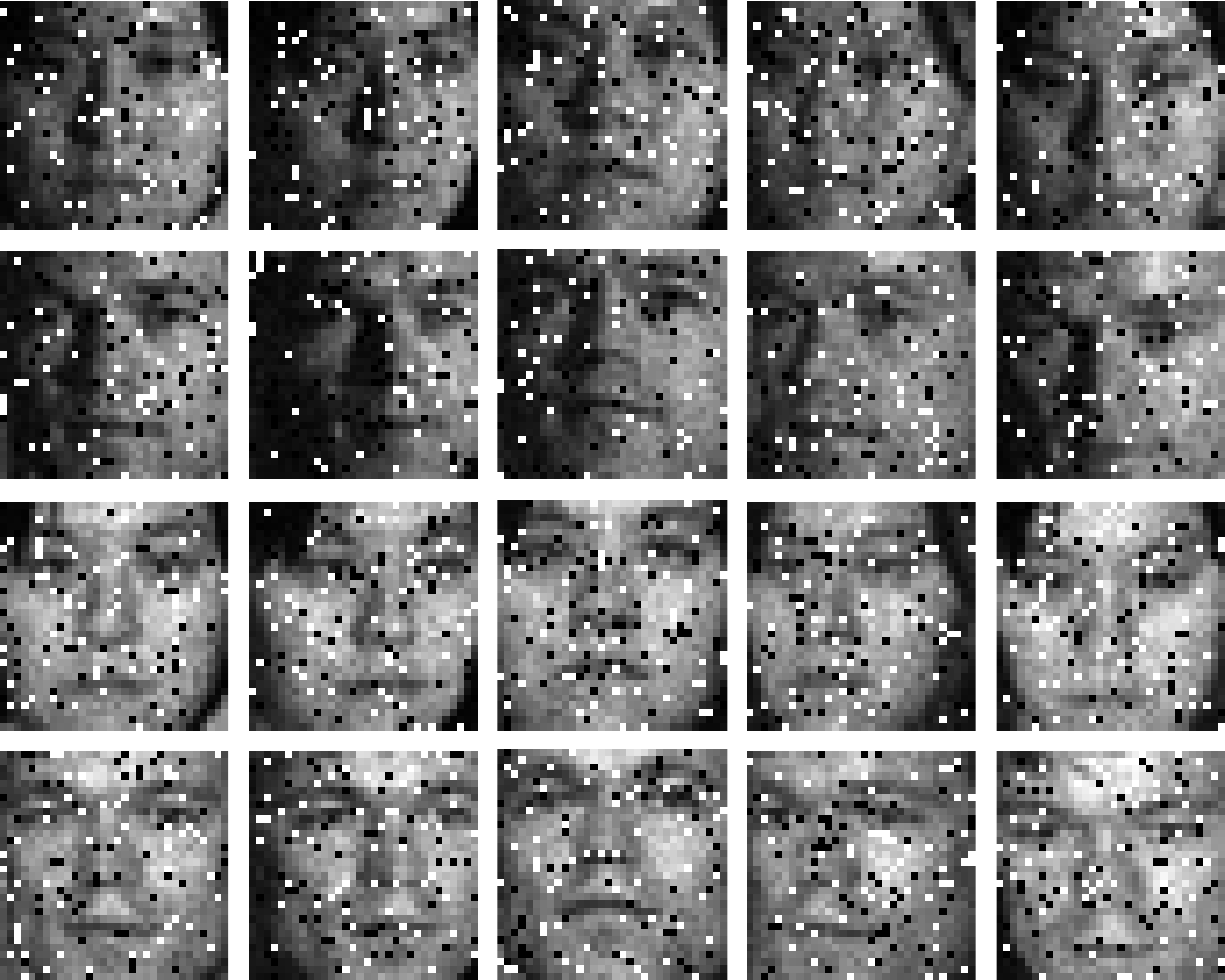}
   \label{fig:face1}
 }
 \subfigure[\small BRTF]{
   \includegraphics[width=0.3\textwidth] {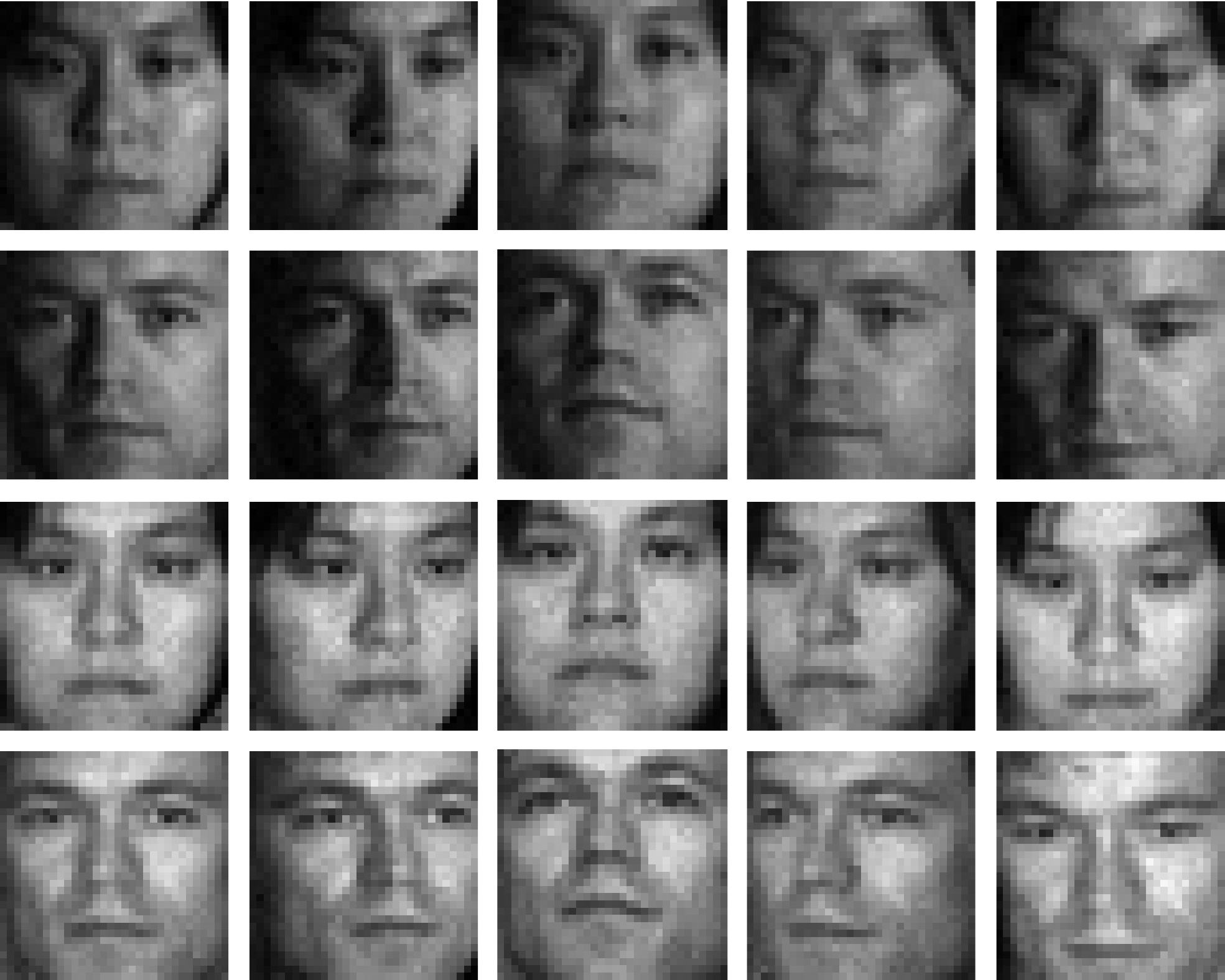}
   \label{fig:face2}
   }
   \subfigure[\small HORPCA]{
   \includegraphics[width=0.3\textwidth] {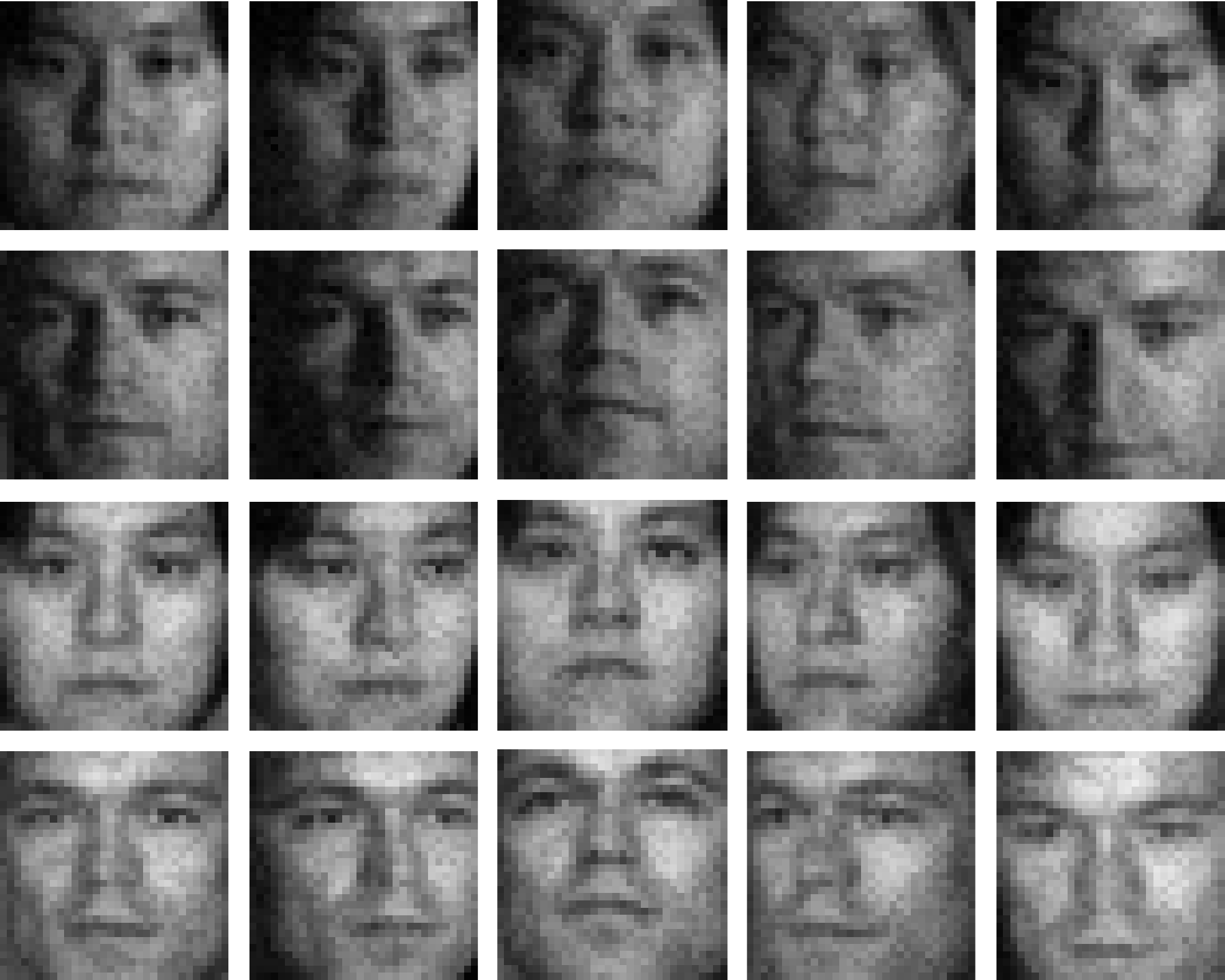}
   \label{fig:face3}
   }
   \subfigure[\small CP-ALS]{
   \includegraphics[width=0.3\textwidth] {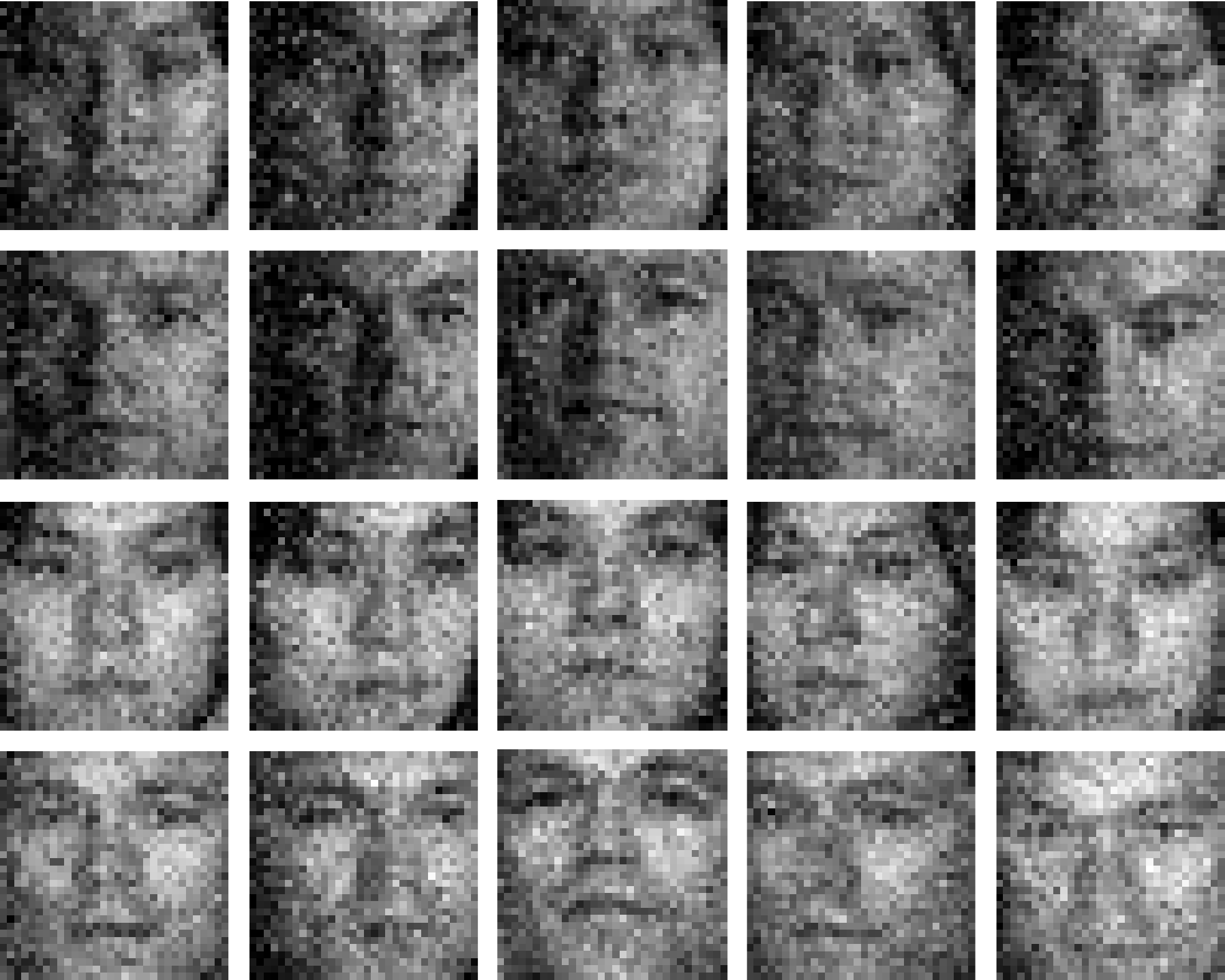}
   \label{fig:face4}
 }
 \subfigure[\small Tucker-ARD]{
   \includegraphics[width=0.3\textwidth] {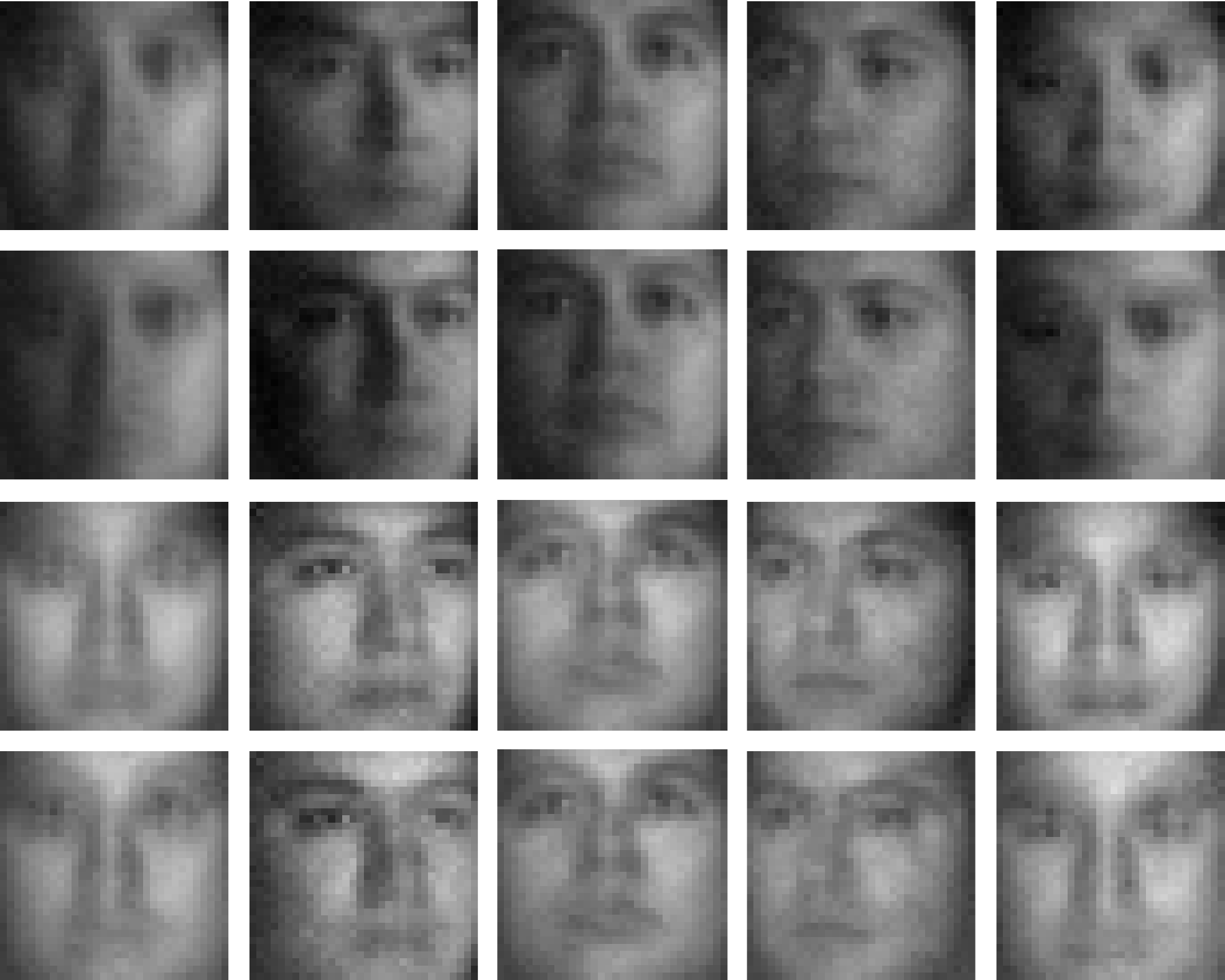}
   \label{fig:face5}
   }
   \subfigure[\small HOSVD]{
   \includegraphics[width=0.3\textwidth] {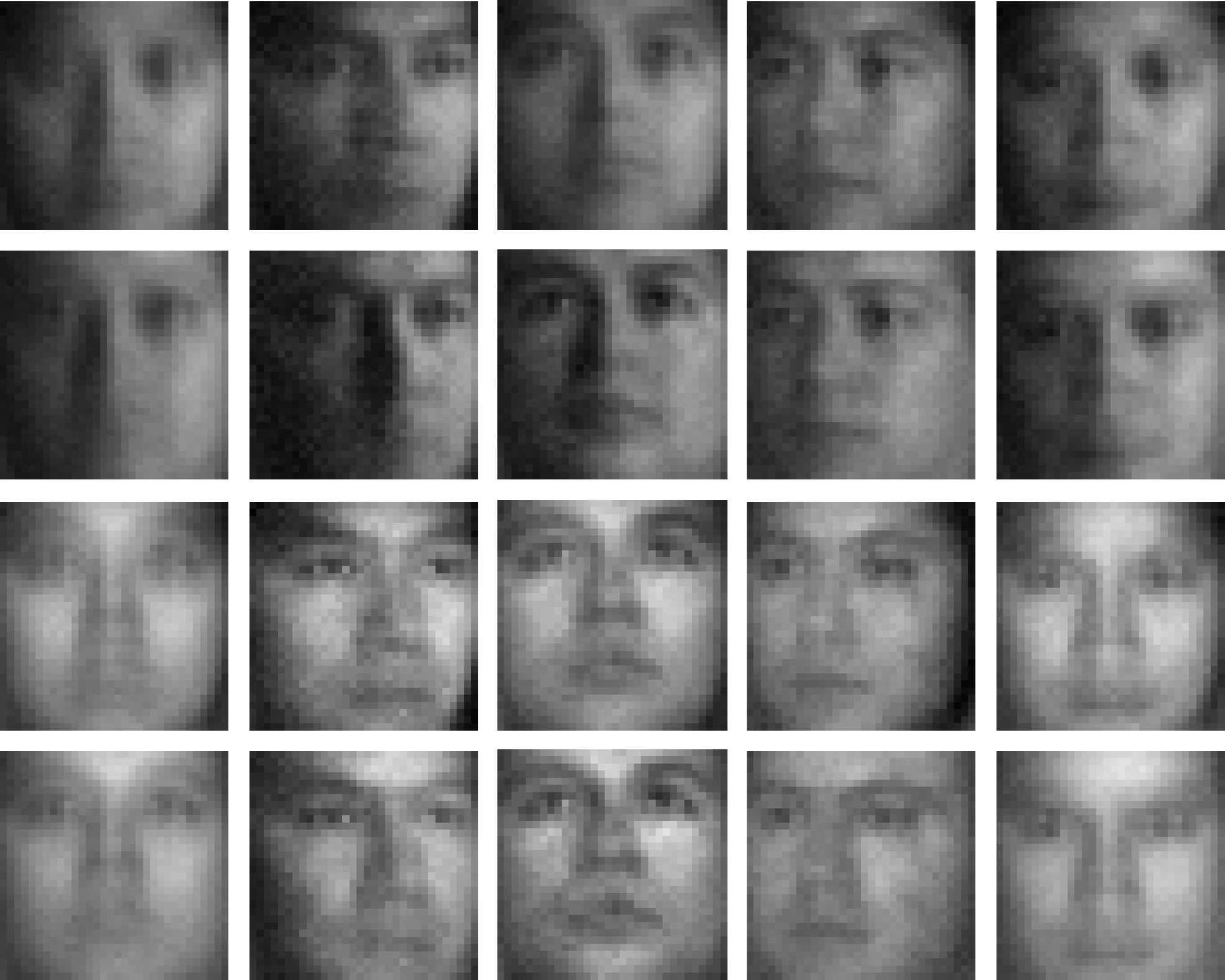}
   \label{fig:face6}
   }
\caption{\subref{fig:face1} The examples of facial images corrupted by poisson and salt-and-pepper noises. \subref{fig:face2}-\subref{fig:face6} The reconstructed images obtained by five tensor factorization and completion methods. }
\label{fig:CMU}
\end{figure*}

In this section, we further investigate the model property in terms of robustness to non-Gaussion noises. We use the CMU-PIE face database~\cite{sim2003cmu} for multilinear model analysis and evaluation. All the facial images are first aligned by their eye position and then cropped to size $32\times 32$. Next, we use 1500 facial images selected from the first 30 subjects with 5 poses and 10 illumination changes to construct a forth-order tensor $\tensor{X}\in\mathbb{R}^{1024\times 30\times 5\times 10}$. Two common types of noise that arises in the images are salt-and-pepper noise (impulse noise) and poisson noise (shot photon noise), which are both non-Gaussian. We consider noise removal for the facial images corrupted by both poisson and salt-and-pepper noise with ratio of 10\% (see Fig.~\ref{fig:CMU}) and compare BRTF with tensor based methods including HORPCA, CP-ALS, Tucker-ARD~\cite{morup2009automatic}, and HOSVD. Note that  the tuning parameters of \mbox{HORPCA} were selected carefully by using the ground-truth of images while BRTF and Tucker-ARD can automatically learn the model parameters without requiring fine-tuning. The rank is initialized as 300 in BRTF, while \emph{CP} rank was finally inferred as 297.  CP-ALS was performed using the $CP$ rank obtained from BRTF and HOSVD was performed using the multilinear rank obtained from Tucker-ARD.

Fig.~\ref{fig:CMU} shows the qualitative results of some example images including 2 people under 5 poses and 2 lighting variations. We can observe that BRTF and HORPCA are robust to non-Gaussian noises and obtain the satisfactory visual quality, while CP-ALS, HOSVD and Tucker-ARD cannot perform well for  non-Gaussian noise removal. The detailed quantitative evaluation results are shown in Table~\ref{tab:CMU} which contains recovery performance evaluated by RRSE and PSNR as well as computational efficiency measured by runtime. BRTF and HORPCA outperform the other methods significantly because of robust property of their model assumptions. It should be noted that although HORPCA tuned parameters by the ground-truth, resulting in the best possible performance, BRTF still outperforms HORPCA in terms of recovery performance. Due to multiple runs for tuning parameter selections, the computation of HORPCA is much more expensive than BRTF. Tucker-ARD is another method that can achieve automatic model selection, however, it cannot handle non-Gaussian noise and its computational efficiency is quite low. These results demonstrate the superiority of BRTF in terms of denoising performance, robustness to non-Gaussian noise and computational efficiency. In summary, BRTF is a robust method which is similar to HORPCA; BRTF can achieve automatic rank determination which is similar to Tucker-ARD; BRTF is based on CP factorization which is similar to CP-ALS. The most significant characteristic of BRTF is the hierarchical probabilistic tensor model and full posterior inference of all unknown variables.

\begin{table}
\renewcommand{\arraystretch}{1.3}
\caption{\small Performance comparisons on facial image denoising. R denotes the inferred rank. }
\label{tab:CMU}
\centering
\resizebox{1\columnwidth}{!}
{
\begin{tabular}{ c | c | c | c | c | c}
 & BRTF & HORPCA & CP-ALS &  Tucker-ARD & HOSVD \\
\hline
 RRSE &  0.0892  & 0.1040   &  0.2551  &  0.2223 &  0.2109 \\
 PSNR &  29.38  & 28.05   &  20.25  &  21.45 &  21.91 \\
 Runtime &  234 s & 1089 s  &  76 s &   4528 s  &  0.7 s \\
 R &  297  & N/A  &  N/A  &   [9, 29, 4, 9]  &  N/A \\
\end{tabular}
}
\vspace{-0.2in}
\end{table}

\section{Conclusion}
\label{sec:conclusion}
In this paper, we have proposed a fully Bayesian generative model for robust tensor factorization, which can naturally handle missing data and outliers, together with the corresponding algorithm for efficient model inference. Our method has several significant characteristics: 1) a general framework for an arbitrary order tensor; 2) robustness to outliers, non-Gaussian noises, and overfitting; 3) tuning parameters free due to an automatic rank determination and automatic parameter selection; 4) the closed-form posterior update for efficient and deterministic inference. Comprehensive experiments and comparisons on synthetic and real-world datasets have confirmed the superiorities of BRTF over  state-of-the-art robust methods and tensor factorization methods. Therefore, BRTF has proven to be promising for robust tensor factorization, robust tensor completion and outlier detection.

\vspace{0.1in}

\emph{\small The Appendix, Matlab codes and demonstration videos are provided in supplementary materials.}

\section*{Acknowledgments}

This work was partially supported by JSPS KAKENHI (Grant Nos 15K15955, 26730125) and the National Natural Science Foundation of China (Grant No. 61202155). The work of Liqing Zhang was supported by the National Natural Science Foundation of China (Grant Nos 61272251, 91120305) and the National Basic Research Program of China (Grant No. 2015CB856004).

\bibliographystyle{IEEEtran}
\bibliography{IEEEabrv,BayesianTensor}

% Generated by IEEEtran.bst, version: 1.12 (2007/01/11)
\begin{thebibliography}{10}
\providecommand{\url}[1]{#1}
\csname url@samestyle\endcsname
\providecommand{\newblock}{\relax}
\providecommand{\bibinfo}[2]{#2}
\providecommand{\BIBentrySTDinterwordspacing}{\spaceskip=0pt\relax}
\providecommand{\BIBentryALTinterwordstretchfactor}{4}
\providecommand{\BIBentryALTinterwordspacing}{\spaceskip=\fontdimen2\font plus
\BIBentryALTinterwordstretchfactor\fontdimen3\font minus
  \fontdimen4\font\relax}
\providecommand{\BIBforeignlanguage}[2]{{%
\expandafter\ifx\csname l@#1\endcsname\relax
\typeout{** WARNING: IEEEtran.bst: No hyphenation pattern has been}%
\typeout{** loaded for the language `#1'. Using the pattern for}%
\typeout{** the default language instead.}%
\else
\language=\csname l@#1\endcsname
\fi
#2}}
\providecommand{\BIBdecl}{\relax}
\BIBdecl

\bibitem{kolda2009tensor}
T.~Kolda and B.~Bader, ``{Tensor Decompositions and Applications},'' \emph{SIAM
  Review}, vol.~51, no.~3, pp. 455--500, 2009.

\bibitem{Cichocki2009}
A.~Cichocki, R.~Zdunek, A.~H. Phan, and S.~I. Amari, \emph{{N}onnegative
  {M}atrix and {T}ensor {F}actorizations}.\hskip 1em plus 0.5em minus
  0.4em\relax John Wiley \& Sons, 2009.

\bibitem{xiong2010temporal}
L.~Xiong, X.~Chen, T.-K. Huang, J.~G. Schneider, and J.~G. Carbonell,
  ``Temporal collaborative filtering with bayesian probabilistic tensor
  factorization,'' in \emph{SDM}, vol.~10.\hskip 1em plus 0.5em minus
  0.4em\relax SIAM, 2010, pp. 211--222.

\bibitem{gao2012probabilistic}
S.~Gao, L.~Denoyer, P.~Gallinari, and J.~GUO, ``Probabilistic latent tensor
  factorization model for link pattern prediction in multi-relational
  networks,'' \emph{The Journal of China Universities of Posts and
  Telecommunications}, vol.~19, pp. 172--181, 2012.

\bibitem{zhao2013higher}
Q.~Zhao, C.~F. Caiafa, D.~P. Mandic, Z.~C. Chao, Y.~Nagasaka, N.~Fujii,
  L.~Zhang, and A.~Cichocki, ``Higher order partial least squares ({HOPLS}): A
  generalized multilinear regression method,'' \emph{{IEEE} Trans. Pattern
  Anal. Mach. Intell.}, vol.~35, no.~7, pp. 1660--1673, Jul. 2013.

\bibitem{chen2013simul}
Y.-L. Chen, C.-T. Hsu, and H.-Y. Liao, ``Simultaneous tensor decomposition and
  completion using factor priors,'' \emph{{IEEE} Trans. Pattern Anal. Mach.
  Intell.}, vol.~36, no.~3, pp. 577--591, Mar. 2014.

\bibitem{de2000multilinear}
L.~De~Lathauwer, B.~De~Moor, and J.~Vandewalle, ``A multilinear singular value
  decomposition,'' \emph{SIAM journal on Matrix Analysis and Applications},
  vol.~21, no.~4, pp. 1253--1278, 2000.

\bibitem{sorensen2012canonical}
M.~S{\o}rensen, L.~D. Lathauwer, P.~Comon, S.~Icart, and L.~Deneire,
  ``Canonical polyadic decomposition with a columnwise orthonormal factor
  matrix,'' \emph{SIAM Journal on Matrix Analysis and Applications}, vol.~33,
  no.~4, pp. 1190--1213, 2012.

\bibitem{acar2011scalable}
E.~Acar, D.~M. Dunlavy, T.~G. Kolda, and M.~M{\o}rup, ``Scalable tensor
  factorizations for incomplete data,'' \emph{Chemometrics and Intelligent
  Laboratory Systems}, vol. 106, no.~1, pp. 41--56, 2011.

\bibitem{sorber2013optimization}
L.~Sorber, M.~Van~Barel, and L.~De~Lathauwer, ``Optimization-based algorithms
  for tensor decompositions: Canonical polyadic decomposition, decomposition in
  rank-({$L_r,L_r,1$}) terms, and a new generalization,'' \emph{SIAM Journal on
  Optimization}, vol.~23, no.~2, pp. 695--720, 2013.

\bibitem{chu2009probabilistic}
W.~Chu and Z.~Ghahramani, ``Probabilistic models for incomplete
  multi-dimensional arrays,'' in \emph{AISTATS Twelfth International Conference
  on Artificial Intelligence and Statistics}, Florida, USA, Apr. 2009, pp.
  89--96.

\bibitem{hayashi2010exponential}
K.~Hayashi, T.~Takenouchi, T.~Shibata, Y.~Kamiya, D.~Kato, K.~Kunieda,
  K.~Yamada, and K.~Ikeda, ``Exponential family tensor factorization for
  missing-values prediction and anomaly detection,'' in \emph{IEEE 10th
  International Conference on Data Mining (ICDM)}, Sydney, NSW, Dec. 2010, pp.
  216--225.

\bibitem{qiinfinite}
Z.~Xu, F.~Yan, and A.~Qi, ``Infinite {Tucker} decomposition: Nonparametric
  {Bayesian} models for multiway data analysis,'' in \emph{Proceedings of the
  29th International Conference on Machine Learning (ICML-12)}, Scotland, UK,
  Jun. 2012, pp. 1023--1030.

\bibitem{elizabeth2013tensor}
E.~S. Allman, P.~D. Jarvis, J.~A. Rhodes, and J.~G. Sumner, ``Tensor rank,
  invariants, inequalities, and applications,'' \emph{SIAM Journal on Matrix
  Analysis and Applications}, vol.~34, no.~3, pp. 1014--1045, 2013.

\bibitem{hillar2013most}
C.~J. Hillar and L.-H. Lim, ``Most tensor problems are {NP-hard},''
  \emph{Journal of the ACM (JACM)}, vol.~60, no.~6, pp. 45:1--45:39, Nov. 2013.

\bibitem{de2008tensor}
V.~De~Silva and L.-H. Lim, ``Tensor rank and the ill-posedness of the best
  low-rank approximation problem,'' \emph{SIAM Journal on Matrix Analysis and
  Applications}, vol.~30, no.~3, pp. 1084--1127, 2008.

\bibitem{alexeev2011tensor}
B.~Alexeev, M.~A. Forbes, and J.~Tsimerman, ``Tensor rank: Some lower and upper
  bounds,'' in \emph{2011 IEEE 26th Annual Conference on Computational
  Complexity}, San Jose, CA, Jun. 2011, pp. 283--291.

\bibitem{burgisser2011geometric}
P.~B{\"u}rgisser and C.~Ikenmeyer, ``Geometric complexity theory and tensor
  rank,'' in \emph{Proceedings of the 43rd annual ACM symposium on Theory of
  computing}, San Jose, California, USA, Jun. 2011, pp. 509--518.

\bibitem{morup2009automatic}
M.~M{\o}rup and L.~K. Hansen, ``Automatic relevance determination for multi-way
  models,'' \emph{Journal of Chemometrics}, vol.~23, no. 7-8, pp. 352--363,
  2009.

\bibitem{rai2014scalable}
P.~Rai, Y.~Wang, S.~Guo, G.~Chen, D.~Dunson, and L.~Carin, ``Scalable bayesian
  low-rank decomposition of incomplete multiway tensors,'' in \emph{Proceedings
  of the 31st International Conference on Machine Learning (ICML-14)}, Beijing,
  CHINA, Jun. 2014, pp. 1800--1808.

\bibitem{liu2013tensorcompletion}
J.~Liu, P.~Musialski, P.~Wonka, and J.~Ye, ``Tensor completion for estimating
  missing values in visual data,'' \emph{{IEEE} Trans. Pattern Anal. Mach.
  Intell.}, vol.~35, no.~1, pp. 208--220, Jun. 2013.

\bibitem{tan2014tensor}
H.~Tan, B.~Cheng, W.~Wang, Y.-J. Zhang, and B.~Ran, ``Tensor completion via a
  multi-linear low-n-rank factorization model,'' \emph{Neurocomputing}, vol.
  133, pp. 161--169, 2014.

\bibitem{signoretto2013learning}
M.~Signoretto, Q.~T. Dinh, L.~De~Lathauwer, and J.~A. Suykens, ``Learning with
  tensors: a framework based on convex optimization and spectral
  regularization,'' \emph{Machine Learning}, vol.~94, no.~3, pp. 303--351,
  2014.

\bibitem{huang2011composite}
J.~Huang, S.~Zhang, H.~Li, and D.~Metaxas, ``Composite splitting algorithms for
  convex optimization,'' \emph{Computer Vision and Image Understanding}, vol.
  115, no.~12, pp. 1610--1622, 2011.

\bibitem{narita2012tensor}
A.~Narita, K.~Hayashi, R.~Tomioka, and H.~Kashima, ``Tensor factorization using
  auxiliary information,'' \emph{Data Mining and Knowledge Discovery}, vol.~25,
  no.~2, pp. 298--324, 2012.

\bibitem{candes2011robust}
E.~J. Cand{\`e}s, X.~Li, Y.~Ma, and J.~Wright, ``Robust principal component
  analysis?'' \emph{Journal of the ACM (JACM)}, vol.~58, no.~3, p.~11, 2011.

\bibitem{xu2012robust}
H.~Xu, C.~Caramanis, and S.~Sanghavi, ``Robust {PCA} via outlier pursuit.''
  \emph{IEEE Transactions on Information Theory}, vol.~58, no.~5, pp.
  3047--3064, 2012.

\bibitem{feng2013online}
J.~Feng, H.~Xu, and S.~Yan, ``Online robust {PCA} via stochastic
  optimization,'' in \emph{Advances in Neural Information Processing Systems},
  Harrahs and Harveys, Lake Tahoe, USA, Dec. 2013, pp. 404--412.

\bibitem{nie2011robust}
F.~Nie, H.~Huang, C.~Ding, D.~Luo, and H.~Wang, ``Robust principal component
  analysis with non-greedy {L1-norm} maximization,'' in \emph{Proceedings of
  the Twenty-Second International Joint Conference on Artificial Intelligence},
  Vancouver, CANADA, Jul. 2011, pp. 1433--1438.

\bibitem{xiong2011direct}
L.~Xiong, X.~Chen, and J.~Schneider, ``Direct robust matrix factorizatoin for
  anomaly detection,'' in \emph{IEEE 11th International Conference on Data
  Mining (ICDM)}, Vancouver, BC, CANADA, Dec. 2011, pp. 844--853.

\bibitem{zheng2012practical}
Y.~Zheng, G.~Liu, S.~Sugimoto, S.~Yan, and M.~Okutomi, ``Practical low-rank
  matrix approximation under robust {$L_1$-norm},'' in \emph{IEEE Conference on
  Computer Vision and Pattern Recognition (CVPR)}, Providence, RI, Jun. 2012,
  pp. 1410--1417.

\bibitem{nie2013joint}
F.~Nie, H.~Wang, H.~Huang, and C.~Ding, ``Joint schatten $p$-norm and
  $l_p$-norm robust matrix completion for missing value recovery,''
  \emph{Knowledge and Information Systems}, vol.~42, no.~3, pp. 525--544, 2013.

\bibitem{Eriksson2012Efficient}
A.~Eriksson and A.~Van~den Hengel, ``Efficient computation of robust weighted
  low-rank matrix approximations using the $l_1$ norm,'' \emph{{IEEE} Trans.
  Pattern Anal. Mach. Intell.}, vol.~34, no.~9, pp. 1681--1690, Sep. 2012.

\bibitem{zhou2011godec}
T.~Zhou and D.~Tao, ``{GoDec:} randomized low-rank \& sparse matrix
  decomposition in noisy case,'' in \emph{Proceedings of the 28th International
  Conference on Machine Learning (ICML)}, Bellevue, Washington, USA, Jun. 2011,
  pp. 33--40.

\bibitem{huang2008robust}
H.~Huang and C.~Ding, ``Robust tensor factorization using {$R_1$} norm,'' in
  \emph{IEEE Conference on Computer Vision and Pattern Recognition (CVPR)},
  Anchorage, AK, Jun. 2008, pp. 1--8.

\bibitem{Zhang2013ICCV}
M.~Zhang and C.~Ding, ``Robust {Tucker} tensor decomposition for effective
  image representation,'' in \emph{The IEEE International Conference on
  Computer Vision (ICCV)}, Sydney, NSW, Dec. 2013, pp. 1550--5499.

\bibitem{li2011robust}
J.~Li, G.~Han, J.~Wen, and X.~Gao, ``Robust tensor subspace learning for
  anomaly detection,'' \emph{International Journal of Machine Learning and
  Cybernetics}, vol.~2, no.~2, pp. 89--98, 2011.

\bibitem{goldfarb2014robust}
D.~Goldfarb and Z.~Qin, ``Robust low-rank tensor recovery: Models and
  algorithms,'' \emph{SIAM Journal on Matrix Analysis and Applications},
  vol.~35, no.~1, pp. 225--253, 2014.

\bibitem{ding2011bayesian}
X.~Ding, L.~He, and L.~Carin, ``Bayesian robust principal component analysis,''
  \emph{{IEEE} Trans. Image Process.}, vol.~20, no.~12, pp. 3419--3430, May
  2011.

\bibitem{luttinen2012bayesian}
J.~Luttinen, A.~Ilin, and J.~Karhunen, ``Bayesian robust {PCA} of incomplete
  data,'' \emph{Neural processing letters}, vol.~36, no.~2, pp. 189--202, 2012.

\bibitem{wang2012probabilistic}
N.~Wang, T.~Yao, J.~Wang, and D.-Y. Yeung, ``A probabilistic approach to robust
  matrix factorization,'' in \emph{Computer Vision--ECCV}.\hskip 1em plus 0.5em
  minus 0.4em\relax Florence, Italy: Springer, Oct. 2012, vol. 7578, pp.
  126--139.

\bibitem{wang2013bayesian}
N.~Wang and D.-Y. Yeung, ``Bayesian robust matrix factorization for image and
  video processing,'' in \emph{Proceedings of International Conference on
  Computer Vision}, Sydney, VIC, Dec., pp. 1785--1792.

\bibitem{babacan2012sparse}
S.~D. Babacan, M.~Luessi, R.~Molina, and A.~K. Katsaggelos, ``Sparse {Bayesian}
  methods for low-rank matrix estimation,'' \emph{{IEEE} Trans. Signal
  Process.}, vol.~60, no.~8, pp. 3964--3977, May 2012.

\bibitem{tipping2001sparse}
M.~E. Tipping, ``Sparse {Bayesian} learning and the relevance vector machine,''
  \emph{The Journal of Machine Learning Research}, vol.~1, pp. 211--244, Sep.
  2001.

\bibitem{winn2005variational}
J.~M. Winn and C.~M. Bishop, ``Variational message passing,'' \emph{Journal of
  Machine Learning Research}, vol.~6, pp. 661--694, Dec. 2005.

\bibitem{zhou2013moving}
X.~Zhou, C.~Yang, and W.~Yu, ``Moving object detection by detecting contiguous
  outliers in the low-rank representation,'' \emph{{IEEE} Trans. Pattern Anal.
  Mach. Intell.}, vol.~35, no.~3, pp. 597--610, Mar. 2013.

\bibitem{sim2003cmu}
T.~Sim, S.~Baker, and M.~Bsat, ``The {CMU} pose, illumination, and expression
  database,'' \emph{{IEEE} Trans. Pattern Anal. Mach. Intell.}, vol.~25,
  no.~12, pp. 1615--1618, Dec. 2003.

\end{thebibliography}

\begin{IEEEbiography}[{\includegraphics[width=1in,height=1.25in,clip,keepaspectratio]{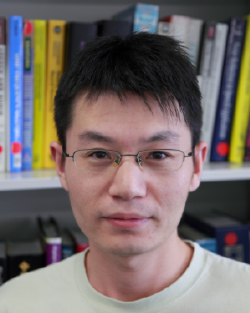}}]{Qibin Zhao} received the Ph.D. degree from Department of Computer Science and Engineering, Shanghai Jiao Tong University, Shanghai, China, in 2009. He is currently a research scientist at Laboratory for Advanced Brain Signal Processing in RIKEN Brain Science Institute, Japan and is also a visiting professor in Saitama Institute of Technology, Japan. His research interests include machine learning, tensor factorization, computer vision and brain computer interface. He has published more than 50 papers in international journals and conferences.
\end{IEEEbiography}

%\vspace{-0.1in}
\begin{IEEEbiography}[{\includegraphics[width=1in,height=1.25in,clip,keepaspectratio]{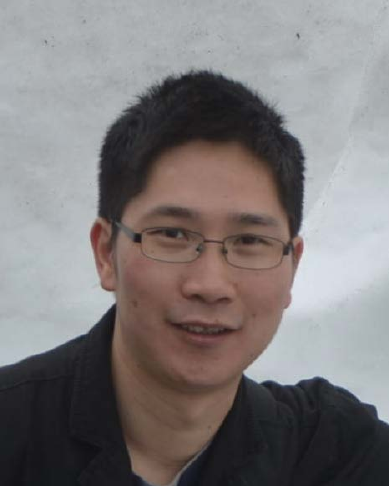}}]{Guoxu Zhou} received the Ph.D degree in intelligent signal and information processing from South China University of Technology, Guangzhou, China, in 2010. He is currently a research scientist of the laboratory for Advanced Brain Signal Processing, at RIKEN Brain Science Institute (JAPAN). His research interests include statistical signal processing, tensor analysis, intelligent information processing, and machine learning.
\end{IEEEbiography}

%\vspace{-0.1in}
\begin{IEEEbiography}[{\includegraphics[width=1in,height=1.25in,clip,keepaspectratio]{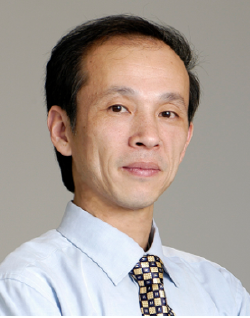}}]{Liqing Zhang} received the Ph.D. degree from Zhongshan University, Guangzhou, China, in 1988. He is now a Professor with Department of Computer Science and Engineering, Shanghai Jiao Tong University, Shanghai, China. His current research interests cover computational theory for cortical networks, visual perception and computational cognition, statistical learning and inference. He has published more than 210 papers in international journals and conferences.
\end{IEEEbiography}

%\vspace{-0.1in}
\begin{IEEEbiography}[{\includegraphics[width=1in,height=1.25in,clip,keepaspectratio]{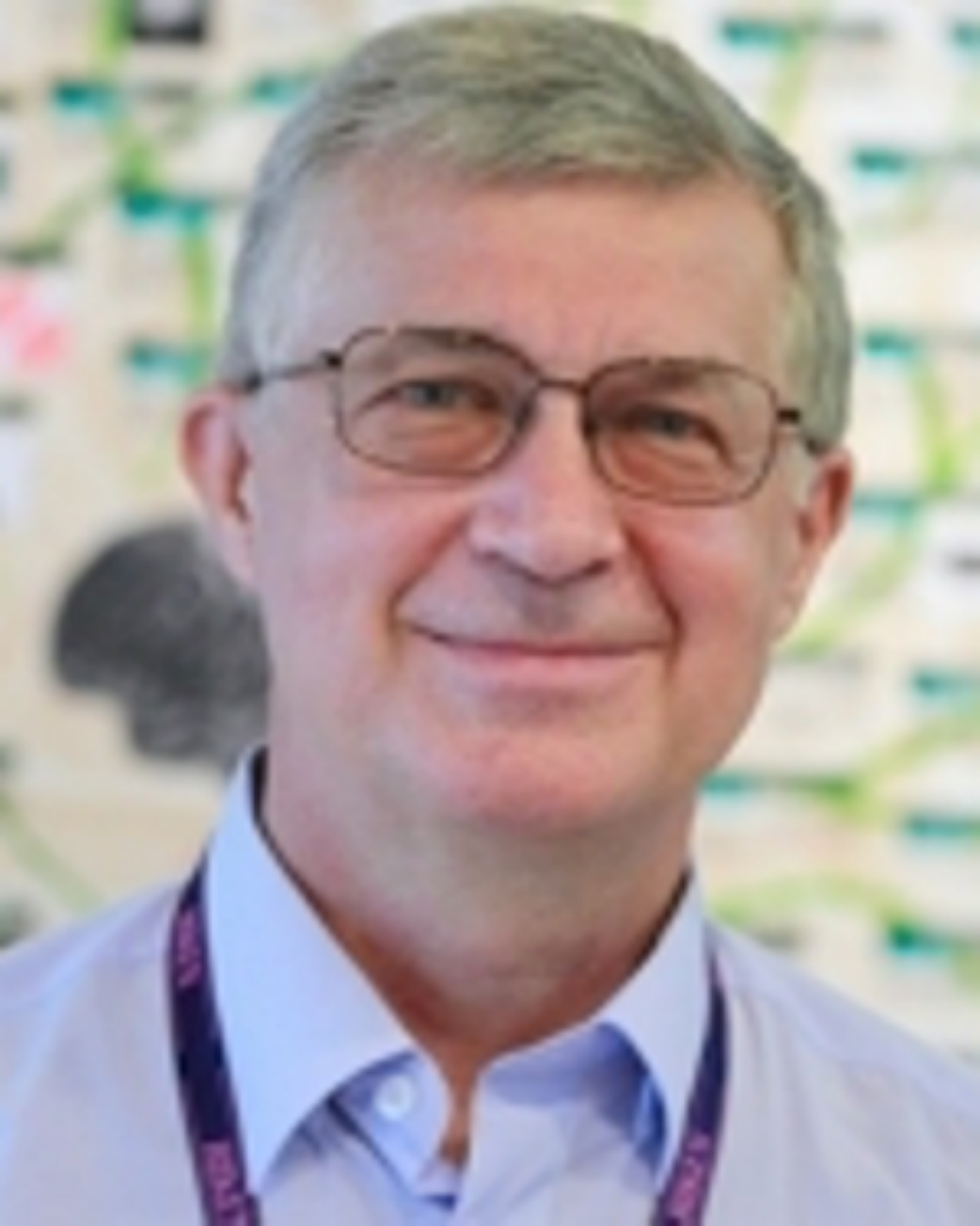}}]{Andrzej Cichocki} received the Ph.D. and Dr.Sc. (Habilitation) degrees, all in electrical engineering, from Warsaw University of Technology (Poland). He is the senior team leader of the Laboratory for Advanced Brain Signal Processing, at RIKEN BSI (Japan). He is coauthor of more than 400 scientific papers and 4 monographs (two of them translated to Chinese). He served as AE of IEEE Trans. on Signal Processing, TNNLS, Cybernetics and J. of Neuroscience Methods.
\end{IEEEbiography}

%\vspace{-0.1in}
\begin{IEEEbiography}[{\includegraphics[width=1in,height=1.25in,clip,keepaspectratio]{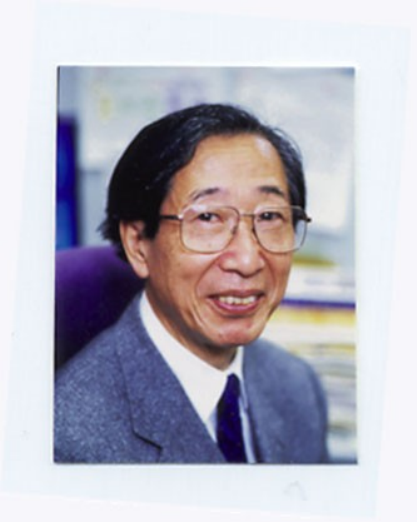}}]{Shun-ichi Amari} received Dr. Eng. degree from the University of Tokyo in 1963.  He had worked at the University of Tokyo and is now Professor-Emeritus.  He served as Director of RIKEN Brain Science Institute and is now its senior advisor.  Dr. Amari served as President of Institute of Electronics, Information and Communication Engineers, Japan and President of International Neural Networks Society.  He received Emanuel A. Piore Award and Neural Networks Pioneer Award from IEEE, the Japan Academy Award, Order of Cultural Merit of Japan, Gabor Award from INNS, among many others.
\end{IEEEbiography}

\vfill

\end{document}